\documentclass{CVM}

\newcommand{\majorrev}[1]{{\color{black}#1}}
\newcommand{\jiamin}[1]{{\color{black}#1}}

\CVMsetup{
type      = {Research/Review Article},
doi       = {s41095-0xx-xxxx-x},
title     = {Hybrid Mesh-neural Representation for 3D Transparent Object Reconstruction},
author    = {Jiamin Xu$^{1,2}$, Zihan Zhu$^{2,3}$, Hujun Bao$^{2}$ and Weiwei Xu$^{2}$\cor{}\\
},
runauthor = {Jiamin Xu, Zihan Zhu, Hujun Bao, Weiwei Xu},
abstract  = {
In this study, we propose a novel method to reconstruct the 3D shapes of transparent objects using images captured by handheld cameras under natural lighting conditions. It combines the advantages of an explicit mesh and multi-layer perceptron (MLP) network as a hybrid representation to simplify the capture settings used in recent studies. After obtaining an initial shape through multi-view silhouettes, we introduced surface-based local MLPs to encode the vertex displacement field (VDF) for reconstructing surface details. The design of local MLPs allowed representation of the VDF in a piecewise manner using two-layer MLP networks to support the optimization algorithm. Defining local MLPs on the surface instead of on the volume also reduced the search space. Such a hybrid representation enabled us to relax the ray-pixel correspondences that represent the light path constraint to our designed ray-cell correspondences, which significantly simplified the implementation of a single-image-based environment-matting algorithm. We evaluated our representation and reconstruction algorithm on several transparent objects based on ground truth models. The experimental results show that our method produces high-quality reconstructions that are superior to those of  state-of-the-art methods using a simplified data-acquisition setup. 
},
keywords  = {Transparent Object, 3D Reconstruction, Environment Matting, Neural Rendering },
copyright = {The Author(s) 2022.},
}

\begin{document}

\maketitle

    \begin{figure}[b] \vskip -4mm
    \small\renewcommand\arraystretch{1.3}
        \begin{tabular}{p{80.5mm}} \toprule\\ \end{tabular}
        \vskip -4.5mm \noindent \setlength{\tabcolsep}{1pt}
        \begin{tabular}{p{3.5mm}p{80mm}}
    $1\quad $ & Hangzhou Dianzi University, Hangzhou, 310018, China. E-mail: Jiamin Xu, superxjm@yeah.net \cor{}.\\
    $2\quad $ & State Key Lab of CAD\&CG, Zhejiang University, Hangzhou, 310058, China.  Jiamin Xu, superxjm@yeah.net; Zihan Zhu, zhuzihan2000@gmail.com; Hujun Bao, bao@cad.zju.edu.cn; Weiwei Xu, xww@cad.zju.edu.cn \cor{}.\\
    $3\quad $ & ETH Zürich, Zürich, 8092, Switzerland. E-mail: Zihan Zhu, zhuzihan2000@gmail.com.\\

&\hspace{-5mm} Manuscript received: 2022-01-01; accepted: 2022-01-01\vspace{-2mm}
    \end{tabular} \vspace {-3mm}
    \end{figure}

\section{Introduction}
\label{sec:intro}
The acquisition of 3D models is a frequent problem in computer graphics and vision. Most existing methods, such as laser scanning and multi-view reconstruction, are based on observations of surface color. Consequently, the surface is assumed to be opaque and approximately Lambertian. These methods cannot be directly applied to transparent objects because the appearance of a transparent object is indirectly observed owing to the complex refraction and reflection light paths at the interface between air and transparent materials.

\begin{figure}[t!]
\centering
\includegraphics[width=1.0\linewidth]{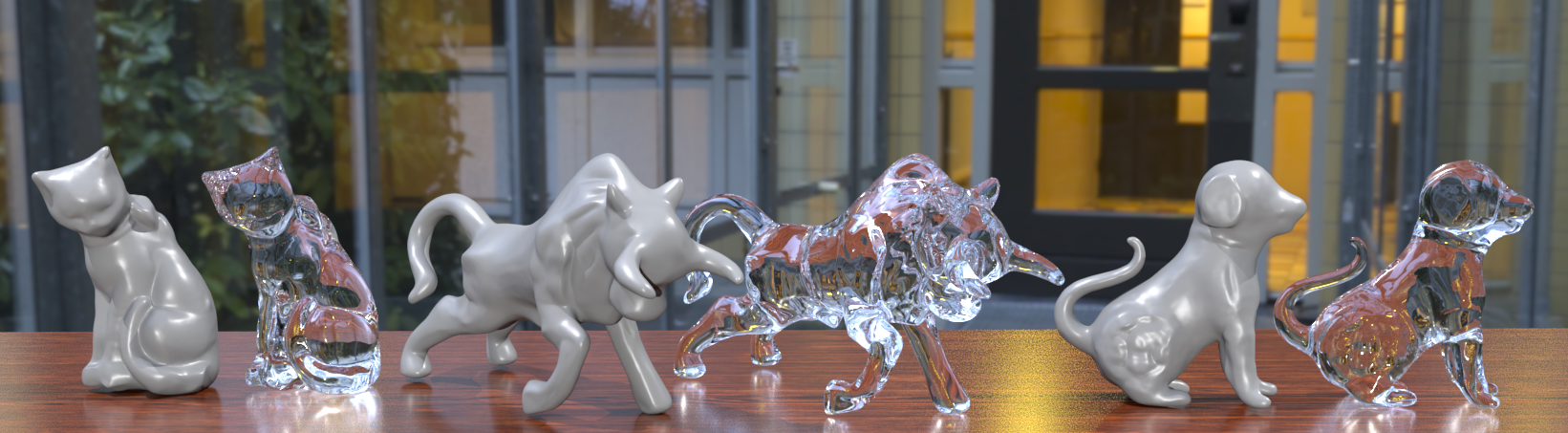}
\caption{Our reconstruction results paired with the associated renderings of three transparent objects. The fine surface details can be reconstructed well via our method using images captured with a handheld camera under natural lighting conditions.} 
\label{fig:teaser} 
\end{figure}

A core technical challenge in 3D transparent object reconstruction is that of handling the dramatic changes in appearance that occur when observing an object in a multi-view setting. Slight changes in an object's shape can lead to nonlocal changes in appearance owing to the complexity of light paths. To address this issue, we utilized ray-pixel correspondence (i.e., the correspondence between a camera ray and a pixel on a static background pattern displayed on a monitor) and ray-ray correspondence (i.e., the correspondence between a camera ray and the incident ray from the background pattern) to provide light path constraints to facilitate 3D transparent object reconstruction~\cite{kutulakos2008theory,wu2018full,ihrke2010transparentsurvey}. A differentiable refraction-tracing technique can be applied to reduce the complexity of the capture setting, and the 3D shape can be recovered through ray-pixel correspondences as shown in~\cite{lyu2020differentiable}. However, in this method, a transparent object should be placed on a turntable under controlled lighting conditions. Li et al.~\cite{li2020through} trained a physics-based neural network to handle complex light paths for 3D transparent objects. The network was trained on a synthetic dataset with a differentiable path tracing rendering technique. This method optimizes surface normals in a latent space; thus, it can reconstruct 3D transparent objects under natural lighting conditions when receiving an environment map and a few images as input. However, this frequently produces overly smooth reconstruction results. 

In this study, we consider how to combine the advantages of explicit meshes and multilayer perceptron (MLP) networks, a hybrid representation, to address the problem of reconstructing transparent objects under natural lighting conditions using images captured with a handheld camera. This representation can be reconstructed through optimization using a differentiable path-tracing rendering technique. The key idea is to use MLP to encode a vertex displacement field (VDF) defined on a base mesh to reconstruct surface details, wherein the base mesh is created using multi-view silhouette images. Our design is motivated by two observations. First, the representation of functions using MLP has been demonstrated to be efficient in optimization and robust to noise~\cite{mildenhall2020nerf,yariv2020multiview,oechsle2021unisurf}. The MLP network parameterizes the VDF with weight parameters globally. Hence, it implicitly provides global constraints on changes in VDF.  Second, defining the MLP-parameterized VDF on the base mesh reduces the search space during optimization~\cite{chen2018deeplightfield}. This significantly accelerates the optimization process compared to MLP-based volumetric representation.

The advantage of our hybrid representation is that it allows for relaxation of the capture setting. Because the global smoothness constraints between vertex displacements are implied in MLP weights, the ray-pixel correspondence required in the optimization can be significantly relaxed to a ray-cell correspondence in our pipeline. Consequently, we can simplify the background pattern design and develop a robust single-image environment matting (EnvMatt) algorithm for handling images captured under natural lighting conditions. Compared to the capture settings used in Wu et al.~\cite{wu2018full} and Lyu et al.~\cite{lyu2020differentiable}’s work, our handheld capture setting is low-cost and simple. Moreover, we propose to represent VDF using a small number of local MLPs. Each MLP is responsible for encoding a local VDF. This strategy enables the design of small-scale MLPs to further accelerate the optimization process. A fusion module is designed to disperse the gradient information of the displacement vectors of vertices to their neighboring local MLPs. This module helps maintain global constraints between local MLPs and produces high-quality reconstruction results. 


The  contributions of this study are summarized as follows.
\setlist{nosep}


\begin{itemize}[leftmargin=14pt]
\item We present a hybrid representation that employs explicit mesh and local-MLP based functions to represent the detailed surface for transparent objects. This approach enables us to design small-scale MLPs to accelerate our optimization algorithm's convergence and achieve high-quality 3D reconstruction results for transparent objects.  
\item We propose a ray-cell correspondence as a relaxed representation of the light path constraint. The ray-cell correspondence is easier to capture, leading to a simplified capture setting under natural lighting conditions. Furthermore, it also eases the implementation of the EnvMatt  algorithm. 
\end{itemize}

The experimental results demonstrate that our method can produce  3D models with details for a variety of transparent objects, as illustrated in Fig.~\ref{fig:teaser}. With our simplified capture setting under natural light conditions, our reconstruction results were superior to those of state-of-the-art 3D reconstruction algorithms for transparent objects.

\section{Related Work}
\label{sec:related}

Our algorithm is designed on the basis of considerable previous research. Here, we review the literature most related to the present work, including studies on transparent object reconstruction, differentiable rendering, and EnvMatt.



\paragraph{Transparent Object Reconstruction} 


Many transparent object reconstruction techniques utilize special hardware setups, including polarization~\cite{cui2017polarimetric,huynh2010shape,miyazaki2005inverse}, time-of-flight cameras ~\cite{tanaka2016recovering}, tomography~\cite{trifonov2006tomographic}, moving point light sources ~\cite{chen2006mesostructure, morris2007reconstructing}, and light field probes\cite{wetzstein2011refractive}. The proposed algorithm is most closely related to shape-from-distortion and light-path triangulation. Kutulakos and Steger~\cite{kutulakos2008theory} formulated the reconstruction of a refractive or mirror-like surface as a light path triangulation problem. Given a function that maps each point
in an image onto a 3D ``reference point'' that is indirectly
projected onto it, the authors characterized a set of reconstructible cases that depended only on the number of points along a light path. The mapping function can be estimated using the EnvMatt algorithm with a calibrated acquisition setup, denoted by ray-point correspondences. A ray-ray correspondence can be uniquely determined with two distinct reference points along the same ray.


In accordance with light path triangulation, one reconstructible case is that of single-refraction surfaces~\cite{schwartzburg2014high,shan2012refractive,yue2014poisson}, particularly fluid surfaces~\cite{morris2011dynamic,qian2017stereo,zhang2014recovering}. Another tractable case is that of transparent objects when rays undergo refraction twice ~\cite{tsai2015does,chari2013theory,qian20163d}. Wu et al.~\cite{wu2018full} recently reconstructed the full shape of a transparent object by first extracting ray-ray correspondences and then performing separate optimization and multi-view fusion. Lyu et al.~\cite{lyu2020differentiable} proposed the extraction of per-view ray-point correspondences using the EnvMatt algorithm in~\cite{zongker1999environment}, and utilized differentiable rendering to progressively optimize an initial mesh. 


In addition to optimization-based methods, deep learning techniques can also be incorporated to resolve depth-normal ambiguities~\cite{stets2019single,sajjan2020clear}. Li et al.~\cite{li2020through} suggested performing optimization in the feature space to obtain surface normals. Subsequently, they performed multi-view feature mapping and 3D point-cloud reconstruction to obtain a 3D shape. Their method works on \jiamin{a simple acquisition setting with only one known environment map and approximately 10 captured images.} However, their reconstructed transparent object may lose some details owing to the domain gap between the real-world images and synthetic training data.

\paragraph{Differentiable Rendering}



In accordance with the simulation level of light transport, differentiable rendering algorithms in computer graphics can be roughly divided into three categories: differentiable
rasterization~\cite{loper2014opendr,kato2018neural,li2018differentiable,liu2019soft,laine2020modular}, differential volumetric rendering~\cite{yariv2020multiview,wang2021neus,yariv2021volume,oechsle2021unisurf}, and differentiable ray-tracing~\cite{nimier2019mitsuba,Li2018DMC,li2015anisotropic,luan2020langevin,zhang2020path,zhang2019differential,bangaru2020unbiased}. Differentiable
rasterization can be used to optimize a mesh itself or its features, and the neural network parameters defined on the mesh. Differentiable volumetric rendering can be used to optimize implicit shape representations, such as the implicit occupancy function~\cite{Mescheder2019OccupancyNetworks,Chen2019LearningImplicit}, signed distance function (SDF)~\cite{Park2019DeepSDF,yariv2020multiview}, and unsigned distance function~\cite{Atzmon2020SAL}. Differentiable rendering has also been used to optimize deep surface light fields~\cite{chen2018deeplightfield}. This method represents per-vertex view-dependent reflections using an MLP. While we also utilize surface-based MLPs, our focus is different; our method employs local MLPs to represent VDF locally to reconstruct surface details and design a fusion layer to avoid discontinuities at the overlapped surface areas.

Considering that a light path with refraction is determined by the front and back surfaces of a transparent object, the geometry can be optimized in an iterative manner with forward ray tracing and backward gradient propagation. To this end, our algorithm exploits differential ray tracing to handle the light path of the reflected and refracted rays on the surface of transparent objects.



\paragraph{Environment Matting}

EnvMatt, which captures how an object refracts and reflects environment light, can be viewed as an extension of alpha matting~\cite{porter1984compositing, levin2007closed}. Image-based refraction and reflection are represented as pixel-texel~(texture pixel) correspondences, in which environments are represented as texture maps. The seminal work of Zongker et al.~\cite{zongker1999environment} extracted EnvMatt from a series of 1D Gray codes, assuming that each pixel is only related to a rectangular background region. Chuang et al. \cite{chuang2000environment} extended this work to recover a more accurate model at the expense of using more structured light backdrops. They also proposed a simplified EnvMatt algorithm that uses only a single backdrop. A pixel-texel correspondence search can also be performed in the wavelet~\cite{peers2003wavelet} and frequency~\cite{qian2015frequency} domains. The number of required patterns can be reduced by combining them with a compressive sensing technique~\cite{duan2015compressive}. Chen et al.~\cite{chen2018tom} recently presented a deep learning framework called TOM-Net to estimate EnvMatt as a refractive flow field. The aforementioned methods require images to be captured under controlled lighting conditions (e.g., in a dark room) to avoid the influence of ambient light. Wexler et al.~\cite{wexler2002image} developed an EnvMatt algorithm for handling natural-scene backgrounds. However, their method required capturing a set of images using a fixed camera and a moving background.


\begin{figure*}[t!]
\centering
\includegraphics[width=1.0\textwidth]{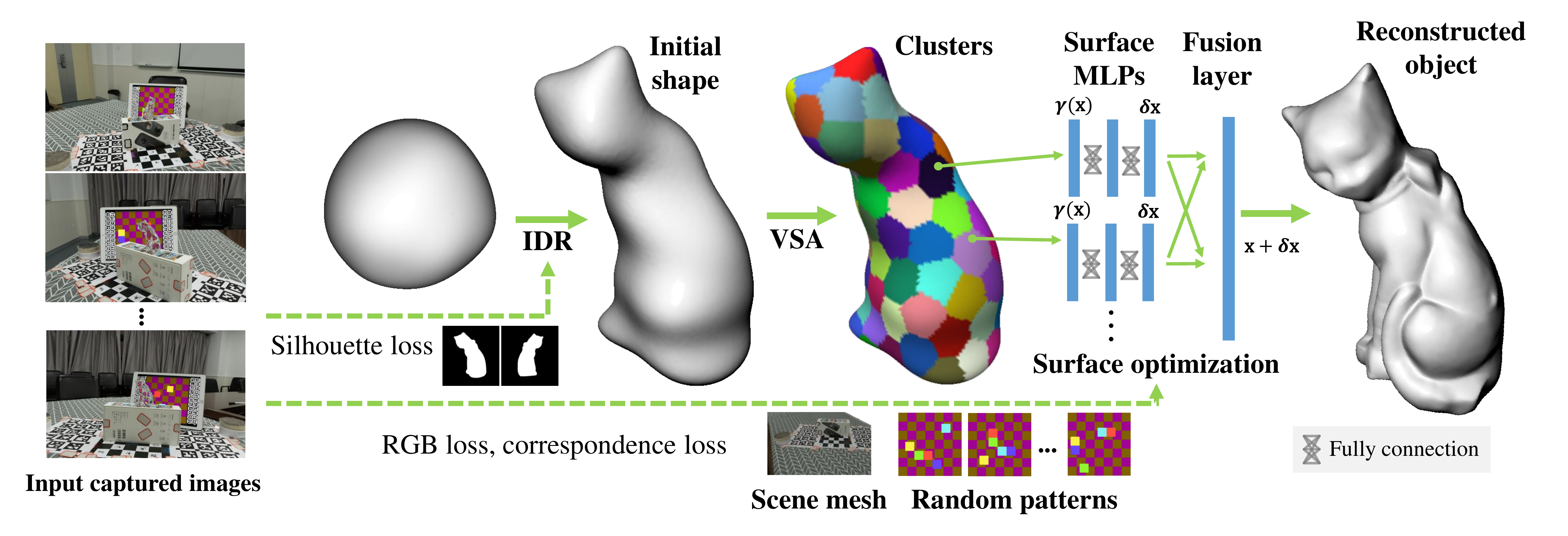}
\caption{The pipeline of our approach.} 
\label{fig:flowchart} 
\end{figure*}

\section{Overview}
\label{sec:overview}

A transparent object reconstruction pipeline is shown in Fig. ~\ref{fig:flowchart}. This pipeline begins by reconstructing an object's rough shape (initial shape) from a collection of multi-view silhouettes. Instead of the space-carving method~\cite{kutulakos2000spacecarving}, we utilized the MLP-based signed distance function (SDF) in IDR~\cite{yariv2020multiview} to obtain a smooth initial shape, as shown in Fig. ~\ref{fig:flowchart}. Subsequently, we employ the MLP network to represent the vertex displacement field (VDF) on the initial shape to reconstruct the surface details. This hybrid surface representation combines an explicit mesh and MLP-based neural network. In the following section, we detail the hybrid representation and optimization algorithm for reconstructing the representation from multi-view images.

\paragraph{Hybrid Representation} We choose to encode the surface details with VDF as it is defined on a 2D manifold instead of the entire 3D space, reducing the search space of the optimization algorithm and producing high-quality reconstruction results. Moreover, we used to represent the displacement field defined on the vertices to simplify the optimization. Such hybrid representation can combine explicit vertex optimization to accelerate convergence and MLP-based neural representation, as in IDR~\cite{yariv2020multiview} to enforce global constraints among vertices and improve the robustness of the optimization.

Rather than encoding the VDF using a single MLP, we found that representing the field with a couple of small local MLPs can achieve better results. As shown in Fig.~\ref{fig:flowchart}, each local MLP encodes the displacement vectors of the vertices within one cluster extracted from the mesh of the initial shape using the variational shape approximation (VSA) algorithm~\cite{cohen2004variational}. To avoid mesh discontinuities across local MLPs, we also added a fusion layer to blend the displacement vectors of neighboring vertices based on the geodesic distances on the mesh~\cite{Crane2017Geodesic}.

\paragraph{Optimization for VDF} The VDF is optimized based on the multi-bounce (up to two-bounces) light path constraints and the consistency between the rendering of our representation and the captured RGB images. The rendering procedure was performed using a recursive differentiable path-tracing algorithm~\cite{li2019differentiable}.

The light path constraint due to multi-bounce refraction is approximated by a mapping function that maps each pixel in the input image onto a pixel of the background pattern image, which can be obtained using the EnvMatt algorithm. We store the background image as a texture. However, we found that traditional EnvMatt algorithms are either restricted to using multiple images with a fixed camera or are sensitive to natural light conditions. Consequently, we designed a grid-based background pattern to establish the correspondence between a foreground pixel $\mathbf{p}$ that covers a small part of the object surface and the cell of the grid. This correspondence is recorded as a tuple $\left< \mathbf{p},\mathbf{u} \right>$, where $\mathbf{u}$ is the pixel coordinate of the cell center on the background image. In this manner, the mapping function is simplified, but remains efficient in providing information on the light path constraint to facilitate optimization. 
~

In the remainder of this paper, we first describe our data pre-processing steps (Sec. ~\ref{sec:pre-processing}), including the image acquisition setup and grid-based single-image EnvMatt algorithm. Then, we present the details of the initial shape reconstruction (Sec. ~\ref{sec:init_shape_recon}) and surface optimization steps (Sec. ~\ref{sec:surface_opt}).

\section{Method}
\label{sec:method}

\subsection{Pre-processing}
\label{sec:pre-processing}

\paragraph{Data Acquisition} We captured images by using a \texttt{Canon} \texttt{EOS} \texttt{60D} digital single-lens reflex camera. The transparent object to be captured was placed on a desk with preprinted AprilTags~\cite{2011AprilTag} underneath. AprilTags were used to facilitate the image registration. To capture the ray-cell correspondences for EnvMatt, we placed an \texttt{iPad} as a monitor behind the transparent object to display a grid-based background pattern. The pattern displayed ${\mathbf{P}_i}$ for the $i$th image was changed after capturing every four images. We manually moved the position of the \ texttt{iPad} after capturing 60 images in our implementation. This setting is designed to incorporate more valid ray-cell correspondences to cover more surface areas. All patterns were pre-generated and used in the EnvMatt step. In addition to images with a background pattern, we captured more images without background patterns to provide further constraints in RGB space. During capture, the \texttt{iPad} was moved two to four times, and the total number of captured images ranged from 154 to 302, owing to the complexity of the object's surface shape.

\begin{figure}[t!]
\centering
\includegraphics[width=1.0\linewidth]{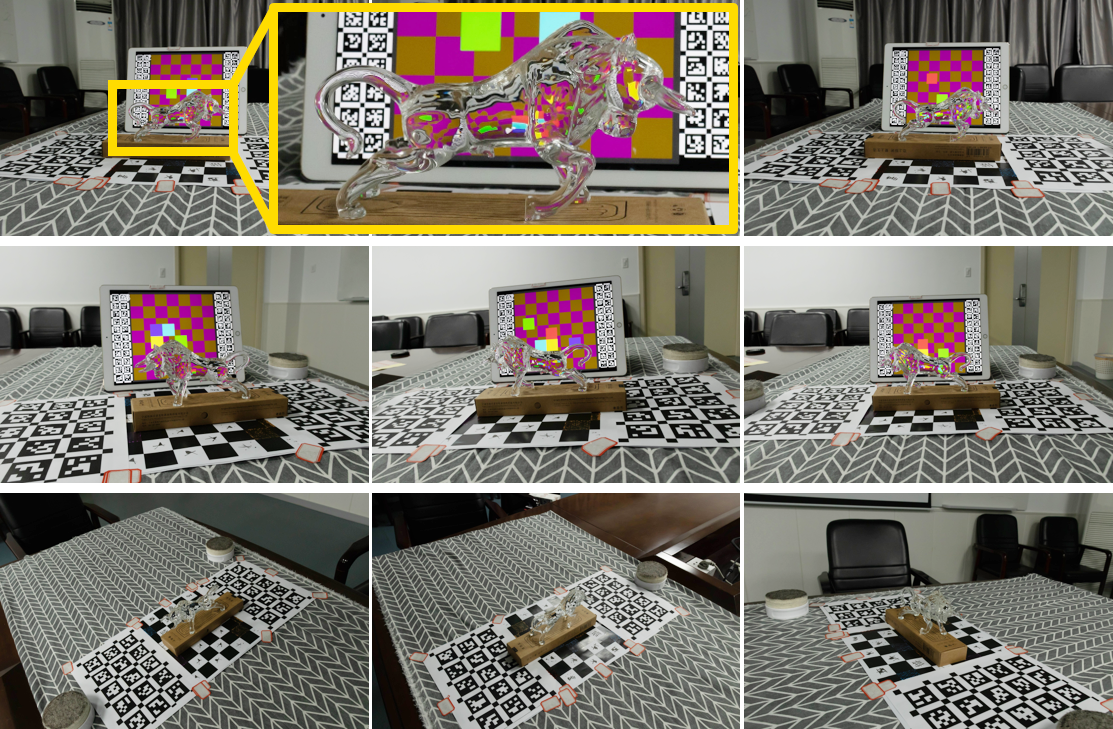}
\caption{Captured images.} 
\label{fig:images} 
\end{figure}



\paragraph{Grid-based Single Image EnvMatt} Similar to \cite{chuang2000environment}, we can assume that a transparent object has no intrinsic color. Therefore, the correspondence between pixels that cover the object's surface and pixels on the background pattern can be calculated by searching a color space. However, we found that the color ramp pattern in~\cite{chuang2000environment} is sensitive to ambient light under natural light conditions because of the invertible and smooth properties of the pattern image in RGB space. Therefore, we designed grid-based background patterns in this study. The color of each grid cell was constant and was designed to create sharp boundaries between cells. As shown in Fig.~\ref{fig:env_matting}, each pattern consists of a $7\times7$ checkboard with two different colors $\{ \mathbf{c}_{h}^{i}|i\in \{ 1,2 \} \}$. Then, we randomly sample five cells on the grid and move each cell with a random offset. These moved cells are assigned one of five distinct salient colors $\{ \mathbf{c}_{s}^{i}|i\in \{ 1..5 \} \}$ as shown in the rightmost column in Fig. ~\ref{fig:env_matting}. We chose these five salient colors because they have sufficient mutual distance in RGB space, which makes them more robust to the influence of environmental lighting. These salient cells are used to calculate the correspondence tuple, and \jiamin{the cell size is chosen as the balance between the surface coverage generated by refracted cells and EnvMatt precision.}

The proposed EnvMatt algorithm calculates ray-cell correspondences for each pixel. Given a captured image, for each pixel $\mathbf{p}$ with color $\mathbf{c}_\mathbf{p}$, the corresponding cell center $\mathbf{u}$ can be calculated as follows.

\begin{footnotesize}
\begin{align}\label{eq:single_img_env_mat}
\mathbf{u}=\begin{cases}
	{cr}( \underset{i}{\text{argmin}}\lVert \mathbf{c}_{s}^{i}-\mathbf{c}_{\mathbf{p}} \rVert ) &		\text{if}\underset{i}{\,\,\min}\lVert \mathbf{c}_{s}^{i}-\mathbf{c}_{\mathbf{p}} \rVert <\gamma_1,\\
	\text{inf} &		\text{if}\underset{i,j}{\,\,\min}( \lVert \mathbf{c}_{h}^{i}-\mathbf{c}_{\mathbf{p}} \rVert ,\lVert \mathbf{c}_{s}^{j}-\mathbf{c}_{\mathbf{p}} \rVert ) >\gamma_2,\\
	\text{none} &		\text{otherwise},
\end{cases}
\end{align}
\end{footnotesize}
where ${cr}(\cdot)$ returns a pre-defined cell center. If $\mathbf{c}_\mathbf{p}$ is similar to any salient color, $\mathbf{u}$ is a valid correspondence. $\mathbf{u}$ is marked as $inf$ when no correspondence exists for $\mathbf{p}$ on the designed grid pattern, indicating that the light path terminates outside the grid. Otherwise, the pixel $\mathbf{p}$ corresponds to a pixel on the grid pattern with checkboard colors. In this case, a precise ray-cell correspondence cannot be obtained. Therefore, we set the correspondence between $\mathbf{p}$ and $\text{none}$. The parameters ${\gamma }_1$ and ${\gamma }_2$ are set as 0.3 and 0.4, respectively.

\jiamin{Once capturing has been performed, we use the 3D reconstruction software \texttt{RealityCapture}~\cite{Realitycapture2016} to register the captured images, which enables us to trace rays in a unified coordinate frame during differentiable rendering. Please refer to sec.~1 in supp. for details.}



\begin{figure}[t!]
\centering
\includegraphics[width=1.0\linewidth]{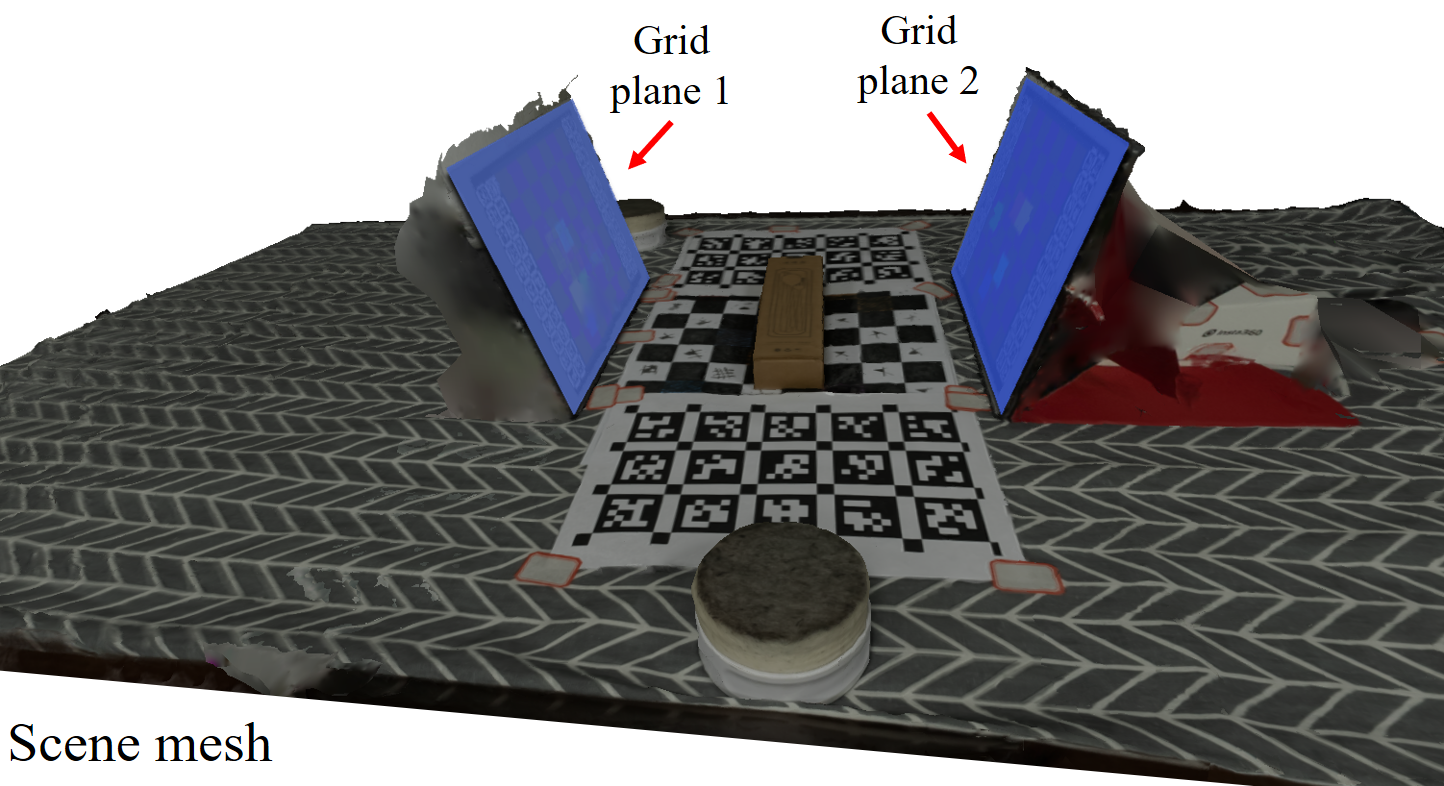}
\caption{Scene mesh and pattern planes. The iPad is moved twice. During image registration step, images are registered to the same global coordinates.} 
\label{fig:scene_mesh} 
\end{figure}

\begin{figure}[t!]
\centering
\includegraphics[width=1.0\linewidth]{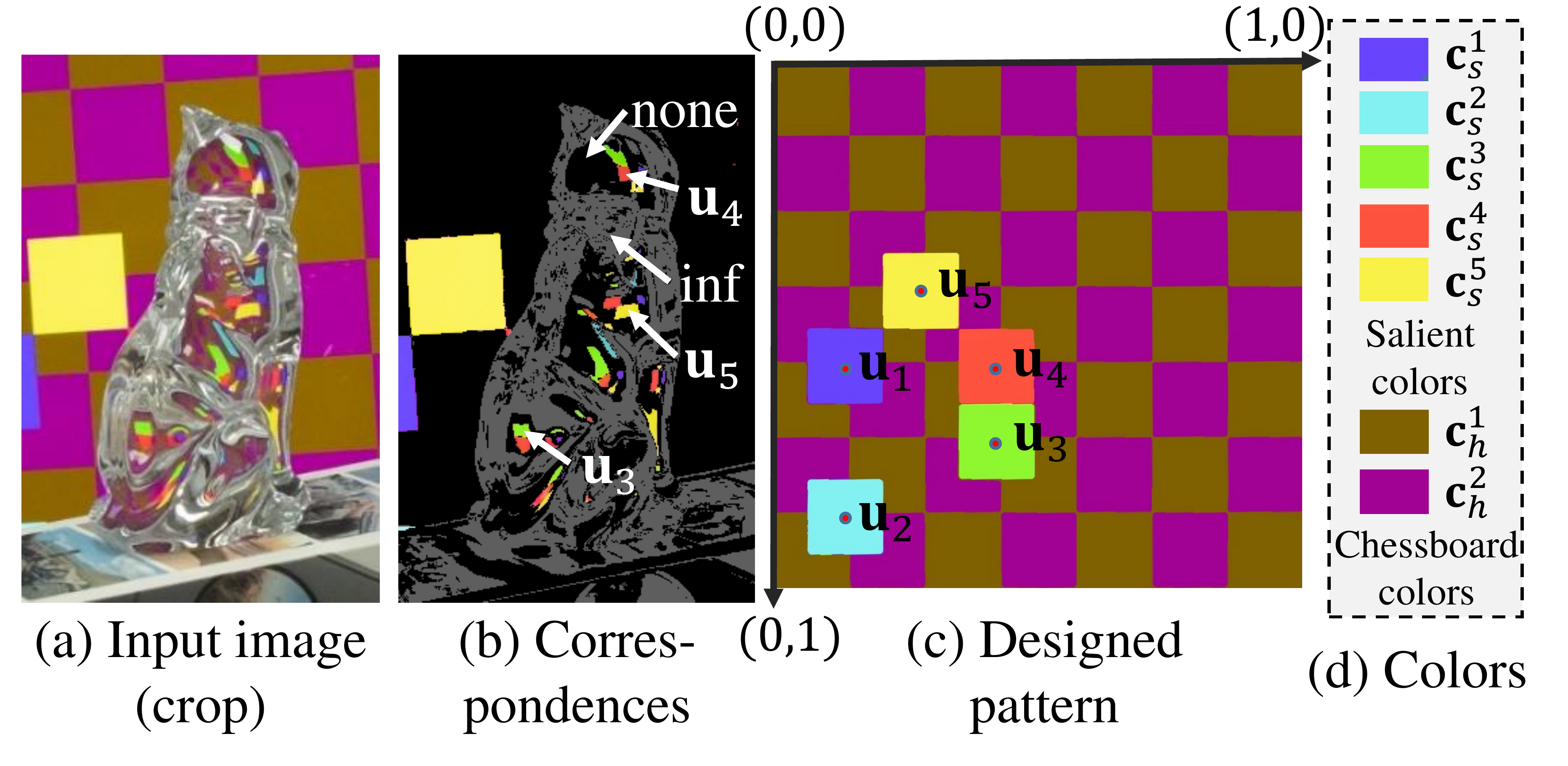}
\caption{Grid-based single image EnvMatt procedure. (a) Input image. (b) EnvMatt results: colored pixels indicate that their traced rays terminate inside the cells with designed salient colors; black pixels indicate that the rays terminate inside cells with checkboard colors, and gray pixels indicates that rays terminate outside the pattern. (c) The designed pattern. The circles indicate the centers of salient cells. (d) Chosen colors for salient and checkboard cells.} 
\label{fig:env_matting} 
\end{figure}

\paragraph{Image Registration and 3D Reconstruction} After capturing the images (please see example images illustrated in Fig.~\ref{fig:images}), we used the 3D reconstruction software RealityCapture~\cite{Realitycapture2016} to register the captured images, enabling us to trace rays in a unified coordinate frame during differentiable rendering. Given that the position of the \texttt{iPad} is changed every 60 images to provide more ray-cell correspondences, we reconstructed the 3D scene with textures that contain the iPad every 60 images independently, resulting in several components in RealityCapture. Each component recorded the information of each independently reconstructed 3D scene. All components are then registered based on AprilTags~\cite{2011AprilTag} beneath the object, as shown in Fig. ~\ref{fig:scene_mesh}. 

Considering that the background pattern displayed on the iPad is changed during the capturing of every four images, incorrect matching points on the iPad's surface are produced, resulting in failure of 3D iPad plane reconstruction. To address this issue, we displayed additional AprilTags surrounding the background patterns to add extra matching points to guarantee the success of the iPad plane reconstruction.

Each 3D plane $\mathbf{P}_i$ of the background pattern was detected using the RANSAC shape-detection algorithm~\cite{Schnabel2010Efficient} as illustrated in Fig. ~\ref{fig:scene_mesh}. We also generated local coordinates for each grid pattern to map the 3D points on the plane onto the texture coordinates. 




\subsection{Initial Shape Reconstruction}
\label{sec:init_shape_recon}

\begin{figure}[t!]
\centering
\includegraphics[width=1.0\linewidth]{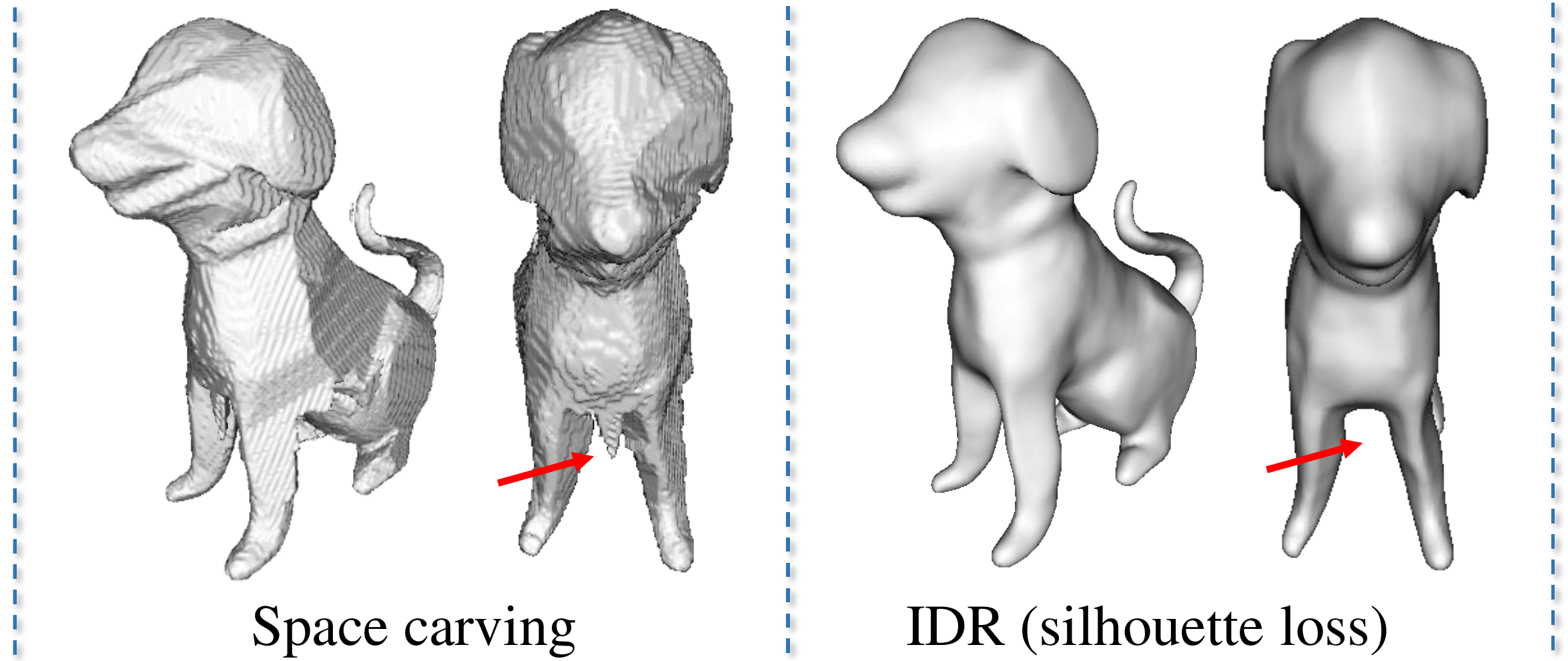}
\caption{Space carving vs. IDR with silhouette loss. The space carving method produces artifact in the occluded areas (indicated by the red arrow).} 
\label{fig:shape_recon_comp} 
\end{figure}

We utilized IDR~\cite{yariv2020multiview} with silhouette (mask) loss to obtain the initial shape of a transparent object. The object masks were manually annotated on several selected images \jiamin{(see Sec. ~4 in supp. for details)}. The number of masks used is listed in Tab.~\ref{table:dataset}. We only used silhouette loss, and thus, the ``neural renderer'' MLP in IDR was removed. As shown in Fig.~\ref{fig:shape_recon_comp}, IDR with silhouette loss produces smoother reconstruction results than the space-carving algorithm. \jiamin{After reconstructing the initial mesh, we uniformly scale the mesh such that the diameter of its bounding ball equals one, extract fine-grain mesh and perform edge collapse mesh simplification. During simplification, we fixed the target-side length of each triangle to 0.005.}

\subsection{Surface Optimization through Differentiable Rendering}
\label{sec:surface_opt}
Given the initial shape as the base mesh, we first group the mesh triangles into several clusters and then assign each cluster to a surface-based MLP. Thus, VDF can be computed as a fusion of the output of the surface-based local MLPs. In particular, for each vertex $\mathbf{x}_i$ within cluster $C_{\mathbf{x}_i}$, each local MLP $\text{MLP}_k$ outputs a displacement vector $\hat{\delta}{\mathbf{x}}_i$ as follows.
\begin{align}\label{eq:displacement_mlp}
\hat{\delta} \mathbf{x}_i=\text{MLP}_{C_{\mathbf{x}_i}}\left( \mathbf{x}_i \right).
\end{align}
To avoid discontinuity at the cluster boundaries, we introduce a differentiable fusion layer to obtain the final displacement vector $\delta{\mathbf{x}}_i$:

\begin{align}\label{eq:fuse_layer}
\delta \mathbf{x}_i=\sum_j{w\left( \mathbf{x}_i,\mathbf{x}_j \right) \cdot \hat{\delta} \mathbf{x}_j},
\end{align}
where $w\left(\mathbf{x}_i,\mathbf{x}_j \right) = \exp(-d\left(\mathbf{x}_i,\mathbf{x}_j \right)/\sigma)$, and $d$ is the geodesic distance between $\mathbf{x}_i$ and $\mathbf{x}_j$, which is calculated using the heat transportation-based method proposed by Crane et al.~\cite{Crane2017Geodesic}. We set the parameter $\sigma$=0.005 in our experiments. 

\begin{figure}[t!]
\centering
\includegraphics[width=1.0\linewidth]{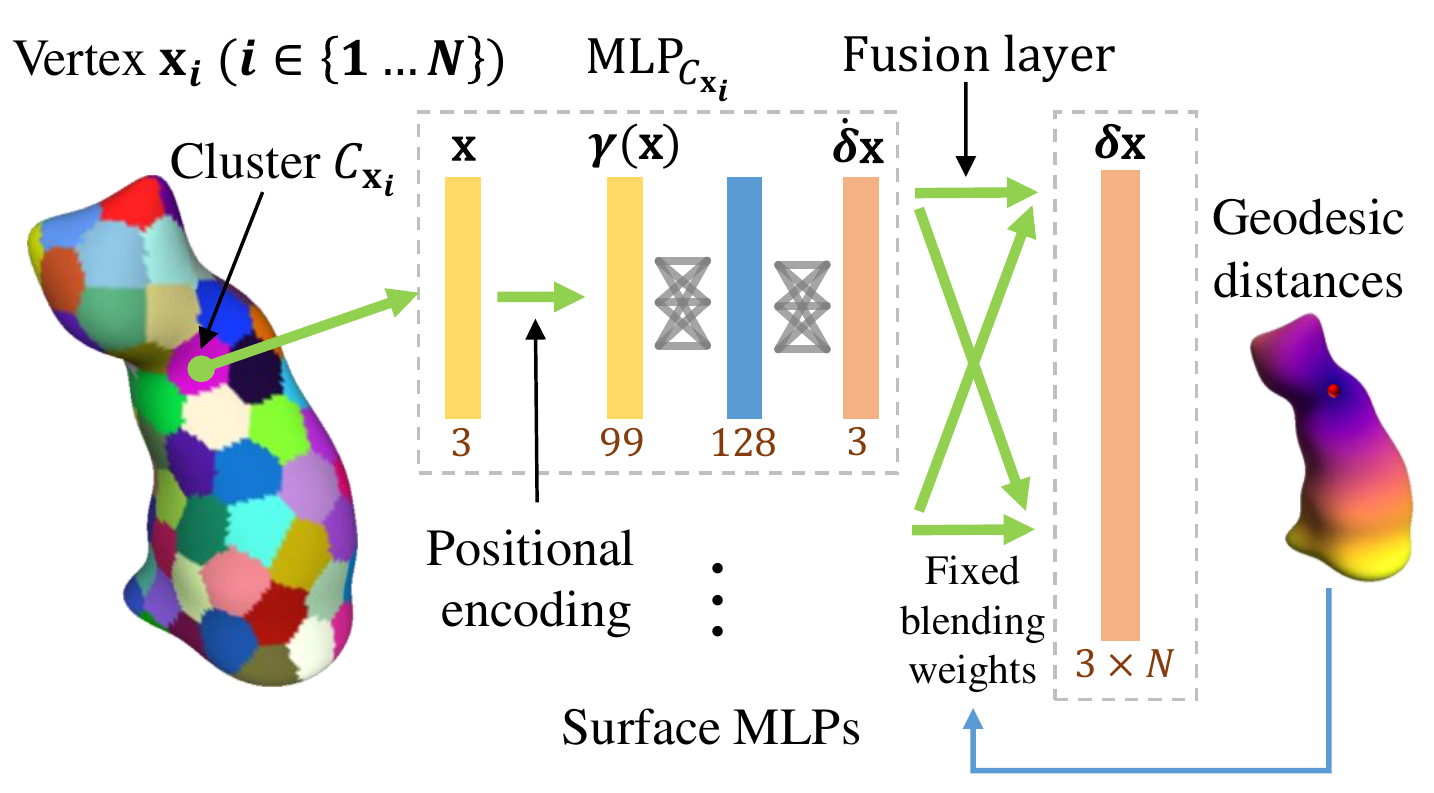}
\caption{Local MLP representation. Each local MLP is responsible for representing the vertex displacements inside a VSA cluster (shown as the colored patch on the object surface). A fusion layer is used to fuse the vertex displacements output by the local MLPs into a smooth VDF on the surface.}
\label{fig:surface_mlps} 
\end{figure}

The architecture of each local MLP is illustrated in Fig. ~\ref{fig:surface_mlps} with two fully connected~(FC) layers. For each MLP, each input 3D vertex on the surface was first mapped to a 99-dimensional feature using positional encoding. We used positional encoding frequencies L=16. In this case, the feature vector size is N=99, where N=L×2×3+3. Here, 2 for sin and cos and 3 for x, y, and z. The first FC layer with weight normalization and ReLu activation maps the 99 dimensional feature to a 128 dimensional latent feature. The second FC layer, with weight normalization and tanh activation, mapped the latent feature to a 3D displacement. Before optimization, all weights were initialized to produce a zero-displacement field.

In the following, we first describe how to extract clusters from the base mesh and then describe the details of the designed loss terms and our optimization procedure. 

\subsubsection{Cluster Extraction}
We utilized the Variational Shape Approximation (VSA) algorithm~\cite{cohen2004variational} to segment the initial shape into several clusters. The VSA algorithm tends to merge co-planar vertices into the same cluster by minimizing $\mathcal{L}^2$ distortion error, which measures the error between a cluster and its linear proxy~(plane). 
To balance the size of each cluster, we add a Euclidean distance error that sums the Euclidean distance between each vertex and its cluster center with a weight of 0.005.




\subsubsection{Loss Terms}

We minimize the following loss function to search for the weight parameters of local MLPs as given below.

\begin{align}\label{eq:total_loss}
\mathcal{L}_{\text{total}}=&\lambda_{\text{rgb}}\mathcal{L}_{\text{rgb}} +\lambda_{\text{corr}}\mathcal{L}_{\text{corr}} + \lambda_{\text{ncorr}}\mathcal{L}_{\text{ncorr}} \\ \notag
+&\lambda_{\text{sil}}\mathcal{L}_{\text{sil}} + \lambda_{\text{reg}}\mathcal{L}_{\text{reg}},
\end{align}
which is the sum of the following five terms: RGB loss $\mathcal{L}_{\text{rgb}}$, ray-cell correspondence loss $\mathcal{L}_{\text{corr}}$, no-correspondence loss $\mathcal{L}_{\text{ncorr}}$, silhouette loss $\mathcal{L}_{\text{sil}}$~\cite{ravi2020pytorch3d}, and regularization loss $\mathcal{L}_{\text{reg}}$. 
The default values of the weights, i.e., $\lambda_{\text{rgb}}$, $\lambda_{\text{corr}}$, $\lambda_{\text{ncorr}}$, $\lambda_{\text{sil}}$ and $\lambda_{\text{reg}}$ were set as $0.001$, $0.1$, $0.03$, $50.0$, and $1.0$, respectively, in our experiments.

\paragraph{RGB Loss} RGB loss measures the difference between the pixel color $\mathbf{c}_\mathbf{p}$ and the rendering result $\mathbf{c}^{in}_\mathbf{p}$ of its traced ray on the basis of the recursive ray tracing algorithm. For a pixel $\mathbf{p}$ on a captured image $\mathbf{I}$, we first trace a ray $\mathbf{l}^{in}_\mathbf{p}$ from the viewpoint associated with $\mathbf{I}$. Then, the recursive ray-tracing algorithm is triggered to  obtain the reflection color $\mathbf{c}^{r1}_\mathbf{p}$ through the first-bounce reflection ray $\mathbf{l}^{r1}_\mathbf{p}$ and the refraction color $\mathbf{c}^{t2}_\mathbf{p}$ through the second-bounce refraction ray $\mathbf{l}^{t2}_\mathbf{p}$ by intersecting and fetching textures from a scene mesh or a corresponding pattern. We consider only single-bounce reflection and double-bounce refraction in the proposed algorithm.

\begin{figure}[t!]
\centering
\includegraphics[width=1.0\linewidth]{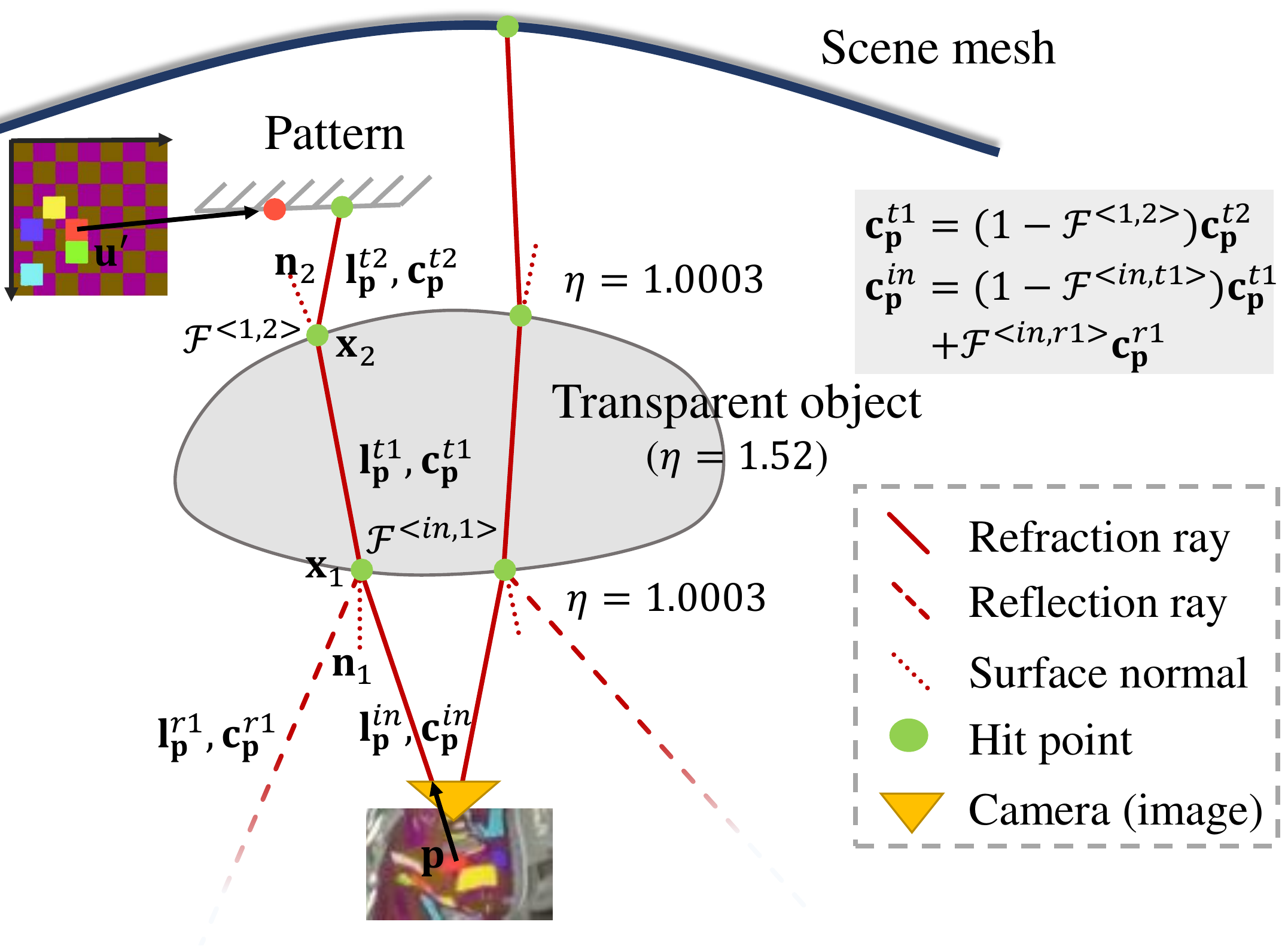}
\caption{Recursive ray-tracing procedure. During rendering, the refraction and reflection rays fetch colors from the scene texture and background pattern. If pixel $\mathbf{p}$ has its corresponding cell center $\mathbf{u}$, then the mesh should be optimized to make the intersection between 
$\mathbf{l}^{t2}_\mathbf{p}$ and the 3D pattern plane within the corresponding cell (the red point in the illustration). } 
\label{fig:reflect_refract} 
\end{figure}

In particular, the camera ray $\mathbf{l}^{in}_\mathbf{p}$ first intersects the front surface at point $\mathbf{x}_1$ and is refracted by the surface with normal $\mathbf{n}_1$ following Snell’s law, as shown in Fig.~\ref{fig:reflect_refract}. \jiamin{The normal of each mesh vertex is obtained by averaging the normals of the triangles incident to the vertex using triangle areas as the weighting shceme, whereas the normal for each ray-triangle intersection point is obtained by interpolating the three vertex normals of the triangle using bary-centric coordinates.} Then, the first-bounce refraction ray $\mathbf{l}^{t1}_\mathbf{p}$ is traced until it hits the back surface at point $\mathbf{x}_2$ with normal $\mathbf{n}_2$. The second-bounce refraction ray $\mathbf{l}^{t2}_\mathbf{p}$ is generated recursively. Similarly, the reflection ray $\mathbf{l}^{r1}_\mathbf{p}$ is traced in the mirror-reflection direction. For RGB loss, we only consider the camera rays with first-bounce reflection ray $\mathbf{l}^{r1}_\mathbf{p}$ and second-bounce refraction ray $\mathbf{l}^{t2}_\mathbf{p}$. We store the valid camera rays whose second-bounce refraction rays can hit the scene mesh or pattern without occlusion and total internal reflection as a binary mask $\mathbf{M}^{t}$ on the image $\mathbf{I}$.

For each refraction ray above, the refraction color is attenuated along the light path, according to the Fresnel term $\mathcal{F}$~\cite{1999Principles}. Taking two successive refraction rays $\mathbf{l}_{\mathbf{p}}^{t1}$ and $\mathbf{l}_{\mathbf{p}}^{t2}$ as an example, where $\mathbf{n}_2$ is the surface normal and $\eta^{i}$ and $\eta^{o}$ are the two indices of refraction (IOR) inside and outside the object, the Fresnel term $\mathcal{F}^{\left< t1,t2 \right>}$ can be computed as follows.


\begin{align}\label{eq:fresnel_term_1}
\mathcal{F}^{\left< t1,t2 \right>}=\frac{1}{2}\left( \frac{\eta ^i\mathbf{r}_{\mathbf{p}}^{t1}\cdot \mathbf{n}_2-\eta ^o\mathbf{r}_{\mathbf{p}}^{t2}\cdot \mathbf{n}_2}{\eta ^i\mathbf{r}_{\mathbf{p}}^{t1}\cdot \mathbf{n}_2+\eta ^o\mathbf{r}_{\mathbf{p}}^{t2}\cdot \mathbf{n}_2} \right) ^2 \\ \notag
+\frac{1}{2}\left( \frac{\eta ^o\mathbf{r}_{\mathbf{p}}^{t1}\cdot \mathbf{n}_2-\eta ^i\mathbf{r}_{\mathbf{p}}^{t2}\cdot \mathbf{n}_2}{\eta ^o\mathbf{r}_{\mathbf{p}}^{t1}\cdot \mathbf{n}_2+\eta ^i\mathbf{r}_{\mathbf{p}}^{t2}\cdot \mathbf{n}_2} \right) ^2, \notag
\end{align}
\begin{align}\label{eq:fresne(fresnel_term_2}
\mathbf{c}_{\mathbf{p}}^{t1}=\left( 1-\mathcal{F}^{\left< t1,t2 \right>} \right) \left( \frac{\eta ^{t1}}{\eta ^{t2}} \right) ^2\mathbf{c}_{\mathbf{p}}^{t2}.
\end{align}
In our experiments, we set the IOR of air to 1.0003 and the IOR of the object material to 1.52. As shown in Fig.~\ref{fig:reflect_refract}, the reflection color $\mathbf{c}^{r1}_\mathbf{p}$ and refraction color $\mathbf{c}^{t2}_\mathbf{p}$ are attenuated after ray tracing, and are used to obtain the rendered color $\mathbf{c}_\mathbf{p}^{in}$.

The reflection and refraction colors are fetched from the textured scene mesh or corresponding pattern. If one reflected ray does not intersect with the scene mesh or pattern, we only set $\mathbf{c}^{r1}_\mathbf{p}$ to zero. RGB loss $\mathcal{L}_{\text{rgb}}$ is defined as follows. 

\begin{align}\label{eq:rgb_loss}
\mathcal{L}_{\text{rgb}}=\frac{1}{\left| \mathbf{M}^t \right|}\sum_{\mathbf{p}}{\mathbf{M}_{\mathbf{p}}^{t}}\lVert \mathbf{c}_{\mathbf{p}}^{in}-\mathbf{c}_{\mathbf{p}} \rVert _1,
\end{align}
where $\lVert\cdot\rVert_1$ indicates the L1 norm, $\mathbf{c}_{\mathbf{p}}^{in}$ is the color rendered from ray tracing, and $\mathbf{c}_{\mathbf{p}}$ is the color of the pixel $\mathbf{p}$. In our implementation, we sampled rays based on $200\times200$ cropped patches instead of individual rays. To match the pixel color and rendered color at the coarse-to-fine level, we filter $\mathbf{c}^{in}_\mathbf{p}$ and $\mathbf{c}_\mathbf{p}$ with three differentiable Gaussian filters~($7\times7, 11\times11, 13\times13$), and then add the outputs together~\cite{zitova2003image}. These Gaussian filters had standard deviations of 2.5, 7.5, and 7.5, respectively. Moreover, we filtered the textures of the scene mesh and patterns using Gaussian filters to smooth the gradients. 




\paragraph{Correspondence Loss and No Correspondence Loss} Correspondence loss $\mathcal{L}_{\text{corr}}$ and no-correspondence loss $\mathcal{L}_{\text{ncorr}}$ were used to enforce the refraction ray $\mathbf{l}^{t2}_\mathbf{p}$ to match the ray-cell correspondence described in Sec.~\ref{sec:pre-processing}. For each pixel $\mathbf{p}$ and its corresponding cell, where the cell center is $\mathbf{u}$ and the side length is $l$, we obtained the ray-plane intersection point between $\mathbf{l}^{t2}_\mathbf{p}$ and the current pattern plane $\mathbf{P}$. Then, the intersection point was projected onto 2D space and transformed into texture coordinates as $\mathbf{v}_{\mathbf{l}_{\mathbf{p}}^{t2}}$. Here, the 2D image space of the planar patterns was pre-rectified to align the texture coordinates with the grid boundaries. As shown in Fig.~\ref{fig:env_matting}, the texture coordinates of the four grid corners were set as $(0, 0)$, $(1, 0)$, $(1, 1)$, and $(0, 1)$, with the texture coordinates of the grid center $g=(0.5, 0.5)$. The correspondence loss $\mathcal{L}_{\text{corr}}$ constrained the projected pixel $\mathbf{v}_{\mathbf{l}_{\mathbf{p}}^{t2}}$ to its corresponding grid center $\mathbf{u}$ as follows.

\begin{align}\label{eq:corr_loss_1}
\mathcal{L}_{\text{corr}}=\frac{1}{\left| \mathbf{M}^t \right|}\sum_{\mathbf{p}}{\mathbf{M}_{\mathbf{p}}^{t}}{d}\left( \mathbf{v}_{\mathbf{l}_{\mathbf{p}}^{t2}},\mathbf{u} \right), 
\end{align}
where $d(\cdot,\cdot)$ is the clipped L2 distance that reduces the loss to zero if the projected pixel is within the cell.
\begin{align}\label{eq:corr_loss_2}
{d}\left( \mathbf{q}_1,\mathbf{q}_2 \right) =\begin{cases}
	\lVert \mathbf{q}_1-\mathbf{q}_2 \rVert _2&		\text{if}~\lVert \mathbf{q}_1-\mathbf{q}_2 \rVert _{\infty}>l/2,\\
	0&		\text{otherwise},\\
\end{cases}
\end{align}
where $l/2$ is the half-side length of a square cell. As we clipped the distance function, our correspondence loss imposed only coarse constraints at the cell level. However, RGB loss can help refine the surface and correspondences at a fine level.

For pixels with no salient correspondence, where $\mathbf{u}=\text{inf}$, we can also add constraints based on the no-correspondence loss $\mathcal{L}_{\text{ncorr}}$. For each pixel with $\mathbf{u}=\text{inf}$, $\mathbf{l}^{t2}_\mathbf{p}$ has no intersection with plane $\mathbf{P}$ or the intersection point is outside the grid area. Thus, the no-correspondence loss $\mathcal{L}_{\text{ncorr}}$ is defined as follows.

\begin{align}\label{eq:no_corr_loss_1}
\mathcal{L}_{\text{ncorr}}=-\frac{1}{\left| \mathbf{M}^t \right|}\sum_{\mathbf{p}}{\mathbf{M}_{\mathbf{p}}^{t}}\hat{d}\left( \mathbf{v}_{\mathbf{l}_{\mathbf{p}}^{t2}},\mathbf{g} \right), 
\end{align}
where the clipped distance $\hat{d}(\cdot,\cdot)$ is the L2 distance when the texture coordinate of the contact point is located outside the grid.
\begin{align}\label{eq:no_corr_loss_2}
\hat{d}\left( \mathbf{q}_1,\mathbf{q}_2 \right) =\begin{cases}
	\lVert \mathbf{q}_1-\mathbf{q}_2 \rVert _2&		\text{if}~\lVert \mathbf{q}_1-\mathbf{q}_2 \rVert _{\infty}<0.5,\\
	0&		\text{otherwise},\\
\end{cases}
\end{align}
where 0.5 is the half-side length of the entire grid pattern in texture coordinates.


\paragraph{Silhouette Loss and Regularization Loss} We also added silhouette loss~\cite{ravi2020pytorch3d} to the annotated object masks similar to the initial shape reconstruction step (Sec.~\ref{sec:init_shape_recon}). Moreover, to further constrain the optimization, we added a regularization loss as follows:

\begin{align}\label{eq:reg_loss}
\mathcal{L}_{\text{\text{reg}}}=\lambda_{\text{ls}}\mathcal{L}_{\text{ls}} + \lambda_{\text{nc}}\mathcal{L}_{\text{nc}} + \lambda_{\text{pc}}\mathcal{L}_{\text{pc}},
\end{align}
which has three loss terms for shape regularization: a Laplacian smoothness loss $\mathcal{L}_{\text{ls}}$ similar to that in \cite{ravi2020pytorch3d}, a normal consistency loss $\mathcal{L}_{\text{nc}}$ as in~\cite{lyu2020differentiable}, and a point cloud regularization loss $\mathcal{L}_{\text{pc}}$ that minimizes the chamfer  distance~\cite{ravi2020pytorch3d} between the optimized points on the current mesh and the initial shape. The losses are defined as follows.

\begin{align}\label{eq:ls_loss}
\mathcal{L}_{\text{ls}}=\sum_{\mathbf{v}_j\in \mathcal{N}\left( \mathbf{v}_i \right)}{\frac{1}{\left| \mathcal{N}\left( \mathbf{v}_i \right) \right|}\left( \mathbf{v}_j-\mathbf{v}_i \right)}, 
\end{align}
\begin{align}\label{eq:nc_loss}
\mathcal{L}_{\text{nc}}=\sum_{e\in \mathcal{E}}{\left(1 - \log \left( 1+\mathbf{n}_{1}^{e}\cdot \mathbf{n}_{2}^{e} \right) \right)},
\end{align}
\begin{align}\label{eq:pc_loss}
\mathcal{L}_{\text{pc}}&=\frac{1}{\left| \mathcal{S}^1 \right|}\sum_{\mathbf{x}^1\in \mathcal{S}^1}{\underset{\mathbf{x}^2\in \mathcal{S}^2}{\min}\lVert \mathbf{x}^1-\mathbf{x}^2 \rVert _{2}^{2}}\\\notag
&+\frac{1}{\left| \mathcal{S}^2 \right|}\sum_{\mathbf{x}^2\in \mathcal{S}^2}{\underset{\mathbf{x}^1\in \mathcal{S}^2}{\min}\lVert \mathbf{x}^1-\mathbf{x}^2 \rVert _{2}^{2}},
\end{align}
where $\mathcal{N}\left( \mathbf{v}_i \right)$ denotes the neighboring vertices of vertex $\mathbf{v}_i$, $\mathcal{E}$ is the set of all edges, and $\mathbf{n}_{1}^{e}\cdot \mathbf{n}_{2}^{e}$ is the dot product of the normals of the two adjacent triangles sharing the same edge $e$. We set $\lambda_{\text{ls}} = 0.2$, $\lambda_{\text{nc}} = 1.0$ and $\lambda_{\text{pc}} = 100.0$.

~

\paragraph{Remark} To increase stability during optimization, we removed three types of camera rays: 1. nearly perpendicular to the surface normals at their intersection with the surface during recursive ray tracing ($\left|\mathbf{l}\cdot\mathbf{n}\right|<0.2$); 2. traced from the pixels near the boundary of the 2D mask created by the shape projection (fewer than seven pixels); and 3. traced from pixels with a nearly overexposed color (larger than 220 in all RGB channels). After these reflection pruning operations, we found that our algorithm can produce high-quality reconstruction results without an environment map, because the remaining rays are dominated by refraction or reflection from the scene mesh.  

Because the light path inside a transparent object is complex, optimizing the surface shape based on local RGB loss was not sufficient, even with our pyramid loss or perceptual loss~(VGG loss). Consequently, we utilized correspondence-based loss to obtain gradients to move $\mathbf{l}^{t2}_\mathbf{p}$ inside the corresponding cell. Within a cell, the distance between a pixel in the rendered image and its corresponding pixel in the captured image was short. Consequently, RGB loss could improve the surface details. We verified the preceding observation through the ablation study discussed in Sec. ~\ref{sec:exp} (with and without correspondence, and no correspondence losses). 

\begin{figure}[t!]
\centering
\includegraphics[width=1.0\linewidth]{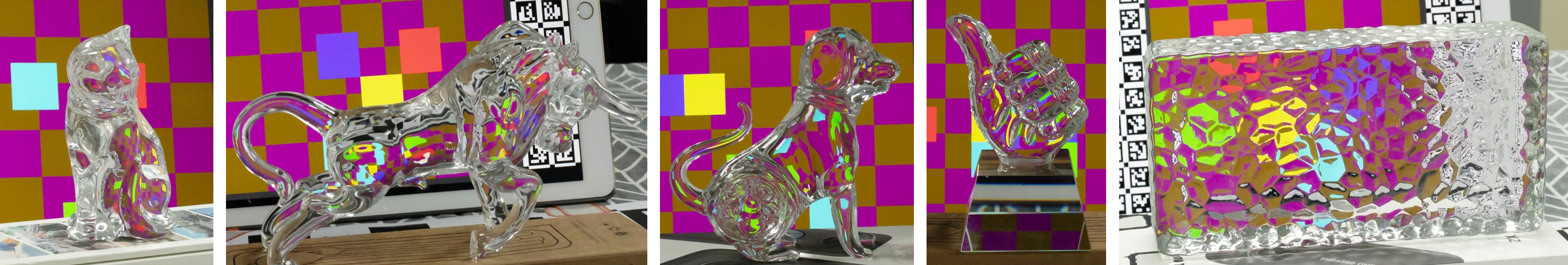}
\caption{Captured images for five transparent objects: the cat object, the cow object, the dog object, the trophy object, and the brick object with bumpy front surface.} 
\label{fig:five_objects_images} 
\end{figure}
\begin{figure}[t!]
\centering
\includegraphics[width=1.0\linewidth]{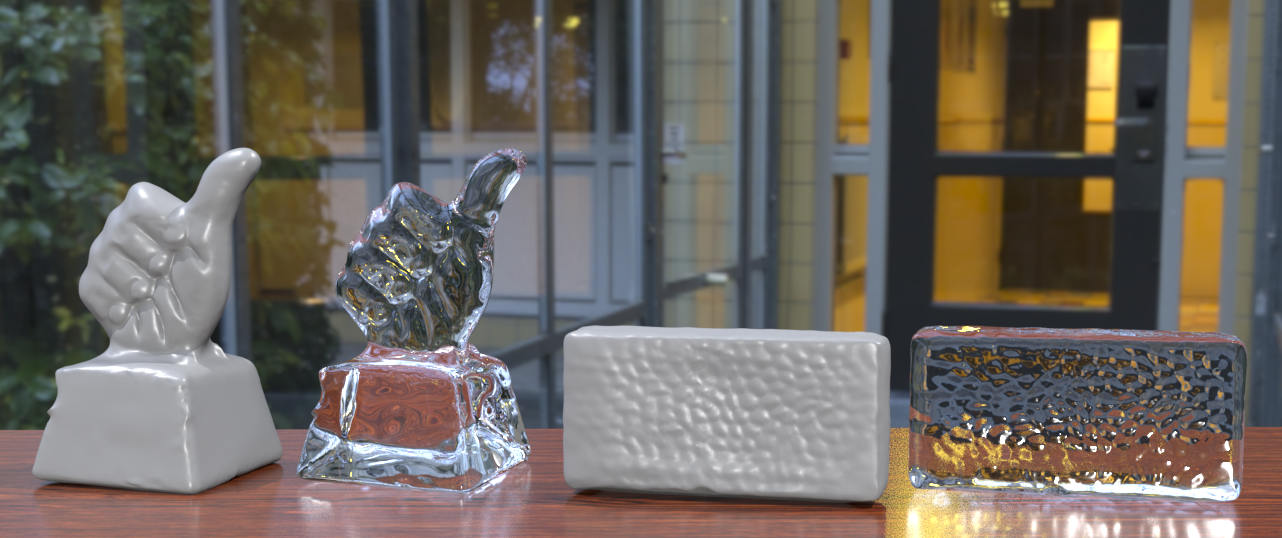}
\caption{The reconstruction results and their corresponding rendering results for a trophy object and a brick object with bumpy front surface.} 
\label{fig:other_objects} 
\end{figure}

\subsection{Implementation Details}
\label{sec:opt}

\paragraph{Initial Shape Optimization} As described earlier, our initial shape reconstruction step is based on IDR~\cite{yariv2020multiview}. We found that with only silhouette loss, an insufficient number of rays may cause holes on the surfaces or sometimes generate another surface beneath the surface of the object. Thus, we increased the number of rays sampled from an image to 20800, and each batch contained rays sampled from three images. We set the learning rate as $1\times10^{-4}$ at the beginning with the same decay strategy as in IDR. The initial shape reconstruction step took approximately 1\textasciitilde2 h after 2000 epochs on NVIDIA GeForce RTX 3090 GPU.

\begin{figure*}[t!]
\centering
\includegraphics[width=0.95\linewidth]{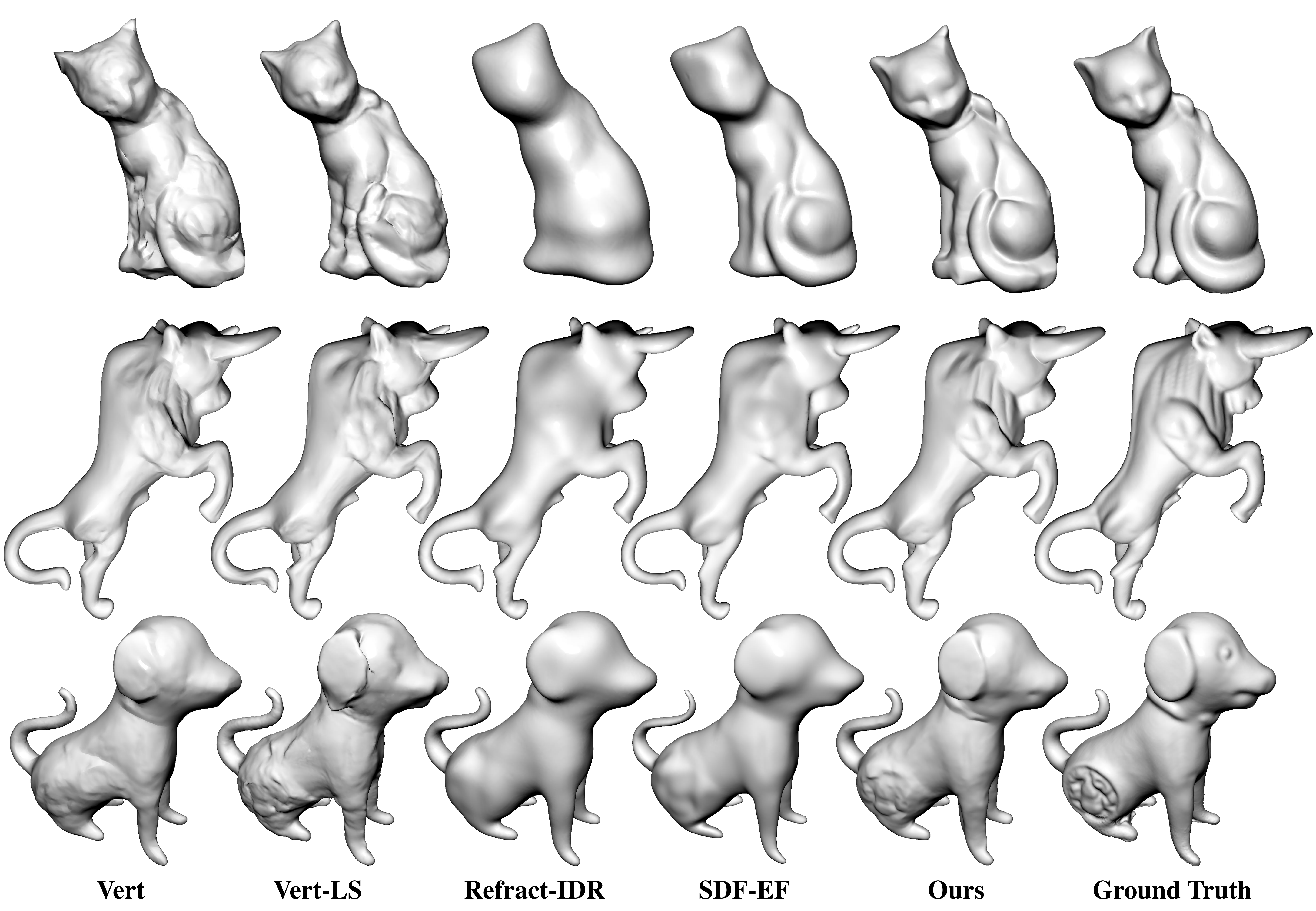}
\caption{Comparisons with
  \textbf{Vert}, \textbf{Vert-LS}{~\cite{nicolet2021large}}, \textbf{Refract-IDR}{~\cite{yariv2020multiview}} and \textbf{SDF-EF}{~\cite{mehta2022level}}.} 
\label{fig:comp_represenataion} 
\end{figure*}

\paragraph{Surface-based MLP Optimization} In the surface-based MLP optimization step, we randomly cropped $200\times200$ image patches from the RGB and silhouette mask images to compute the RGB and silhouette loss terms. In our implementation, we projected the initial shape onto all views to generate the projected mask images, denoted by $\mathbf{M}^{p}$. During optimization, the RGB image patches were sampled if they overlapped with $\mathbf{M}^{p}$ to improve patch sampling efficiency. For silhouette loss, we only sampled crops from masks annotated for the initial shape reconstruction at each iteration. The number of images with annotated masks was much smaller than the total number of captured RGB images in our implementation to ease the burden of the annotation procedure. The optimizer for surface-based local MLPs optimization was ADAM~\cite{Kingma2015AdamAM}, and the learning rate was set to $1\times10^{-5}$ with a cosine annealing~\cite{loshchilov2016sgdr} scheduler. The number of epochs used to train the network was 300. \jiamin{The optimization procedure implemented using PyTorch~\cite{paszke2017automatic} took about one hour on one NVIDIA GeForce RTX 3090 GPU. The ray tracing procedure was accelerated using OptiX engine. For the index of refraction (IOR) coefficient, we fixed the coefficient $\eta$ of the object as 1.52 for all three channels and ignored the dispersion.}


\paragraph{View Selection} \jiamin{We captured images and moved the iPad positions following the condition that for each high-curvature region not near silhouettes, the salient cells in the background pattern were refracted by this region more than twice among all images. Because we did not initially have the object geometry, the high-curvature regions were determined through our observation. The masks were adaptively selected for manual annotation using a view-selection algorithm. This iteratively added views with at least a 60-degree angle distance compared with all previously selected views. Subsequently, some extra views were needed in some cases, as determined by checking the visual hull.}

\section{Experiments}
\label{sec:exp}

We applied our algorithm to reconstruct the 3D shapes of five transparent objects, as shown in Fig. ~\ref{fig:teaser} and Fig. ~\ref{fig:other_objects}, which were made from glass or crystals. The captured images of the five objects are shown in Fig. ~\ref{fig:five_objects_images}. The size of each object, the number of input images, the number of manually annotated masks, and the number of random patterns with moving frequency are listed in Tab. ~\ref{table:dataset}. In the following section, we demonstrate the advantages of our surface-based local MLP representation, perform ablation studies, and compare our method with state-of-the-art transparent object reconstruction methods.


To evaluate the accuracy of reconstruction quantitatively, we painted each object with DPT-5 developer as in \cite{wu2018full} and then scanned it with a scanner to obtain a ground truth mesh in SI unit~(meter), as shown in Fig. ~\ref{fig:comp_represenataion}. We compared the reconstructed results with the ground truth after aligning them using the ICP~\cite{Zhou2018}. Subsequently, we evaluated the reconstruction by measuring the chamfer distance between the two point clouds.

\begin{table}[t!] 
\caption{Statistics of data acquisition. \#Img denotes the number of captured images. \#Mask, \#Pattern, and \#Move denote the number of annotated masks, the number of our grid-based patterns presented on iPad, and the number of iPad movements, respectively.}
\centering
\small
\resizebox{\linewidth}{!}{
\begin{tabular}{|c|c|c|c|c|c|}
\hline
{Object}  & {Size~($cm^3$)} & {\#Img } & {\#Mask} & {\#Pattern} & {\#Move} \\ 
\hline
{cow} & {$9.5*4.5*17.5$}  & {154} & {19} & {30} & {2}\\ 
\hline
{cat} & {$8*2.5*4.5$}  & {302} & {29} & {60} & {4}\\ 
\hline
{dog} & {$11*3.5*11.5$}  & {300} & {19} & {60} & {4}\\ 
\hline
{trophy} & {$12*7*7$}  & {172} & {14} & {30} & {2}\\ 
\hline
{brick} & {$10*4.5*20$}  & {178} & {15} & {30} & {2}\\ 
\hline
\end{tabular}
}
\label{table:dataset}
\end{table}

\begin{table}[t!] 
\caption{The chamfer distance measurement for Fig.~\ref{fig:comp_represenataion}.}
\centering
\small
\resizebox{\linewidth}{!}{
\begin{tabular}{|c|c|c|c|c|c|}
\hline
{Object} & {Vert} & {Vert-LS} & {Refract-IDR} & SDF-EF & Ours \\ 
\hline
cat & {1.010} & {0.792} & {1.397} & {0.978} & \textbf{0.590} \\
\hline
cow & {1.035} & {1.032} & {1.092} & {1.182} & \textbf{0.986}\\ 
\hline
dog & {1.568} & {1.398} & {1.472} & {1.423} & \textbf{1.014}\\
\hline
\end{tabular}
}
\label{tab:comp_represenataion}
\end{table}



\subsection{Evaluations}

\subsubsection{Surface-based Local MLP Representation}

We performed a comparison to verify the importance of the surface-based local MLP representation. Thus, four other shape representations were incorporated: 1. explicit mesh vertices, as in \cite{lyu2020differentiable}, denoted by \textbf{Vert}; 2. mesh vertices with an advanced optimizer in \cite{nicolet2021large}, denoted by \textbf{Vert-LS} (Large Step); 3. SDF encoded by a single MLP, similar to IDR~\cite{yariv2020multiview}, denoted by \textbf{Refract-IDR}; and 4. SDF encoded by a single MLP with explicit flows, as in \cite{mehta2022level}, denoted by \textbf{SDF-EF} (explicit flows). All representations were optimized using the same loss functions in Sec. ~\ref{sec:surface_opt}. The implementation details of the representations used in the comparison are as follows.

\noindent{\textbf{Vert}}: This representation explicitly optimizes the position of each vertex using the same loss as ours and with an ADAM optimizer. The learning rate was set to $1\times10^{-5}$ with a cosine annealing~\cite{loshchilov2016sgdr} scheduler.

\noindent{\textbf{Vert-LS}}: This representation is similar to \textbf{Vert} but with the gradient calculation method and the new optimizer proposed in \cite{nicolet2021large}. The gradient steps in \cite{nicolet2021large} already involve the Laplacian energy. Therefore, we removed the explicit Laplacian smoothness loss. We set the diffusion weight $\lambda$ to 10 and the learning rate to $2\times10^{-3}$ as in their study.

\noindent{\textbf{Refract-IDR}}: \jiamin{The Refract-IDR representation is a modified version of IDR~\cite{yariv2020multiview}. We keep the mask loss and eikonal loss the same as IDR and extend the spherical ray tracing in IDR for two-bounce refract and one-bounce reflection rays. The RGB loss in IDR was replaced by our RGB loss. We also added our corr and ncorr loss, where the weight for each loss term was the same as in our study. During optimization, the gradient of the normal for each intersection point was back-propagated to the MLP through the autograd operation. The architecture of the Refract-IDR network involves eight hidden layers, with each layer having dimensions of 256. The activation function, level of positional encoding, optimizer, and learning rate were set the same as in IDR.}

\noindent{\textbf{SDF-EF}}: \jiamin{The SDF-EF representation is the same as in Mehta et al.~\cite{mehta2022level}. It also utilizes the SDF encoded by a single MLP to represent the surfaces. Unlike the Refract-IDR representation, it back-propagates the gradients to the MLP network based on explicit mesh vertices extracted using the marching cube algorithm~\cite{remelli2020meshsdf,mehta2022level}. During the optimization, this representation performs a marching cube to extract the mesh at each iteration. The architecture of the SDF-MLP network, activation function, level of positional encoding, optimizer, and learning rate were set the same as in Refract-IDR. We also add the Laplacian smooth term, as in \cite{mehta2022level} to regularize the surface.}

\begin{figure}[t!]
\centering
\includegraphics[width=1.0\linewidth]{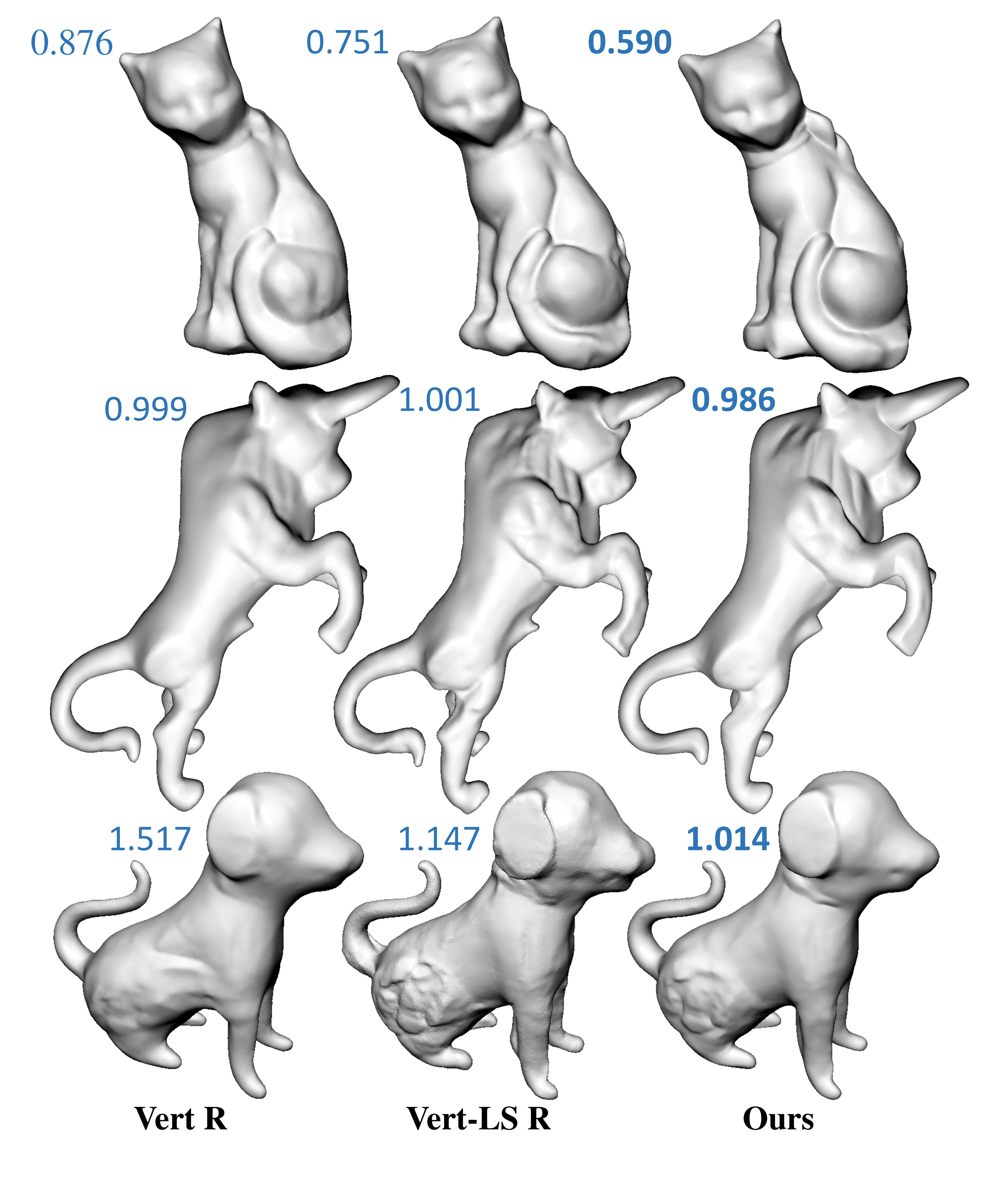}
\caption{Comparisons with \textbf{Vert R} and \textbf{Vert-LS  R}{~\cite{nicolet2021large}}. The numbers indicate the chamfer distances between the mesh and the ground truth illustrated in Fig.~\ref{fig:comp_represenataion}.} 
\label{fig:comp_w_remeshing} 
\end{figure}

As shown in Fig.~\ref{fig:comp_represenataion} and Table. ~\ref{tab:comp_represenataion}, our Surf-MLP representation outperformed other representations. Explicit mesh optimization introduces high-frequency artifacts. Although the Vert-LS method can reduce artifacts, it introduces folds into some areas. Simultaneously, the Refract-IDR and SDF-EF representations led to slow convergence and produced over-smooth results that lost some details.

\majorrev{The explicit mesh optimization described above yields some artifacts because explicit optimization is sensitive to noise. However, these artifacts can be reduced by optimizing the
vertices in a coarse-to-fine manner. We can progressively
performed remeshing during explicit mesh optimization. We
denote Vert and Vert-LS with remeshing as Vert R and Vert
LS R and compared our methods with them. Similar to Lyu et
al. [4], we do remeshing after every fixed iteration.
(30 epochs were used in the experiment). At each remeshing stage: 
The target triangle size is fixed and decreases iteratively. As
shown in Fig. 12, the remeshing procedure can increase 
quality of the reconstructed results compared with Vert and
Vert-LS. However, our method still outperformed these results.
of these two methods based on the chamfer distance.
between the mesh and the ground truth, as shown in Fig. 12.
We believe this is because the weights of the loss terms after each
remeshing stage must be tuned during optimization. If the
weights are not set adaptively, the results for Vert R and the
Vert-LS R may suffer from oversmoothing or local
high-frequency artifact.}




\subsubsection{Comparison with Li et al.~\cite{li2020through} and Lyu et al.~\cite{lyu2020differentiable}}

\begin{figure}[t!]
\centering
\includegraphics[width=1.0\linewidth]{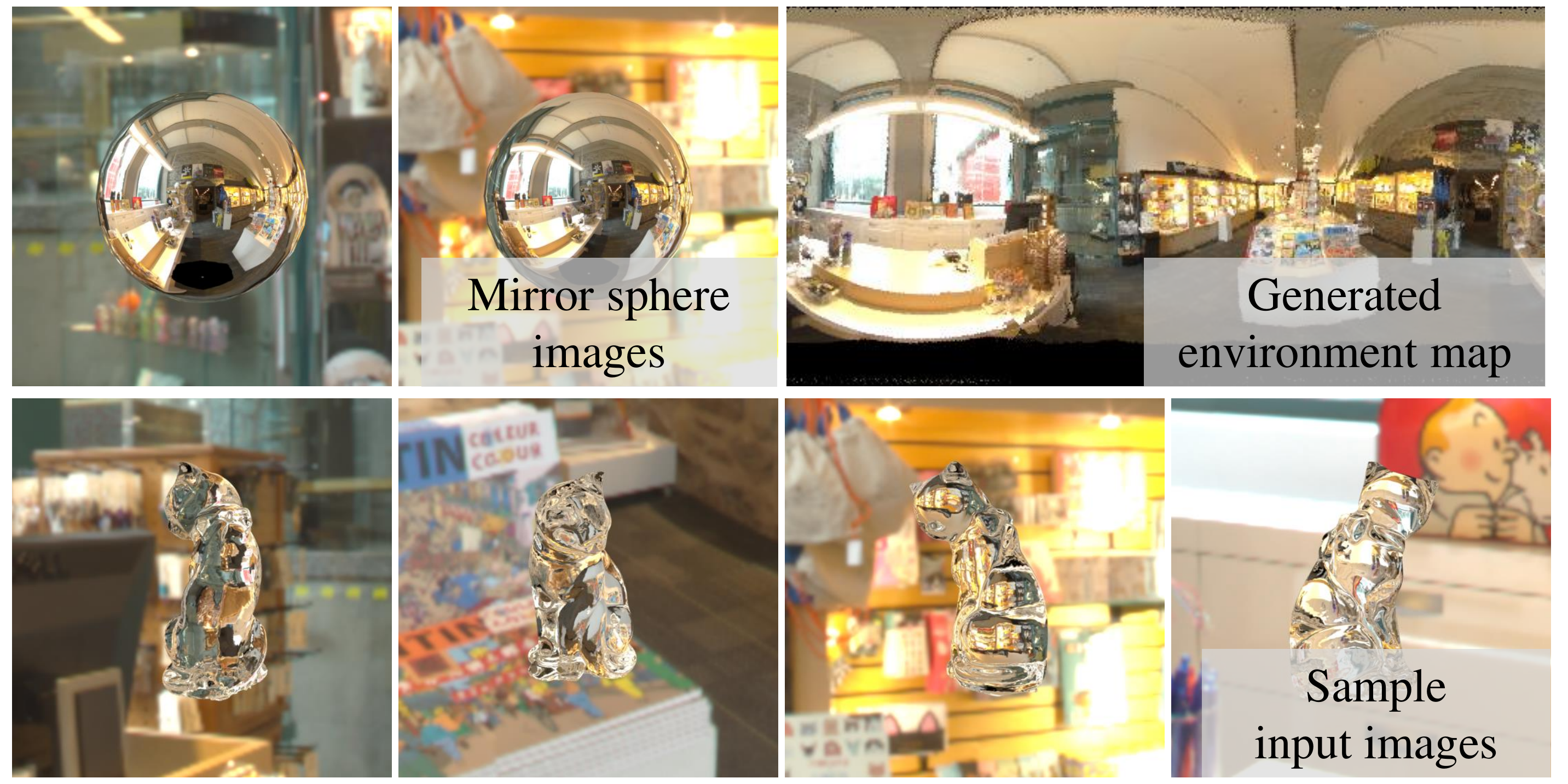}
\caption{The input images for Li et al.~\cite{li2020through}. The mirror sphere images are used in Li et al.~\cite{li2020through} to generate the environment map.}
\label{fig:li_etal_input_images} 
\end{figure}

\begin{figure}[t!]
\centering
\includegraphics[width=1.0\linewidth]{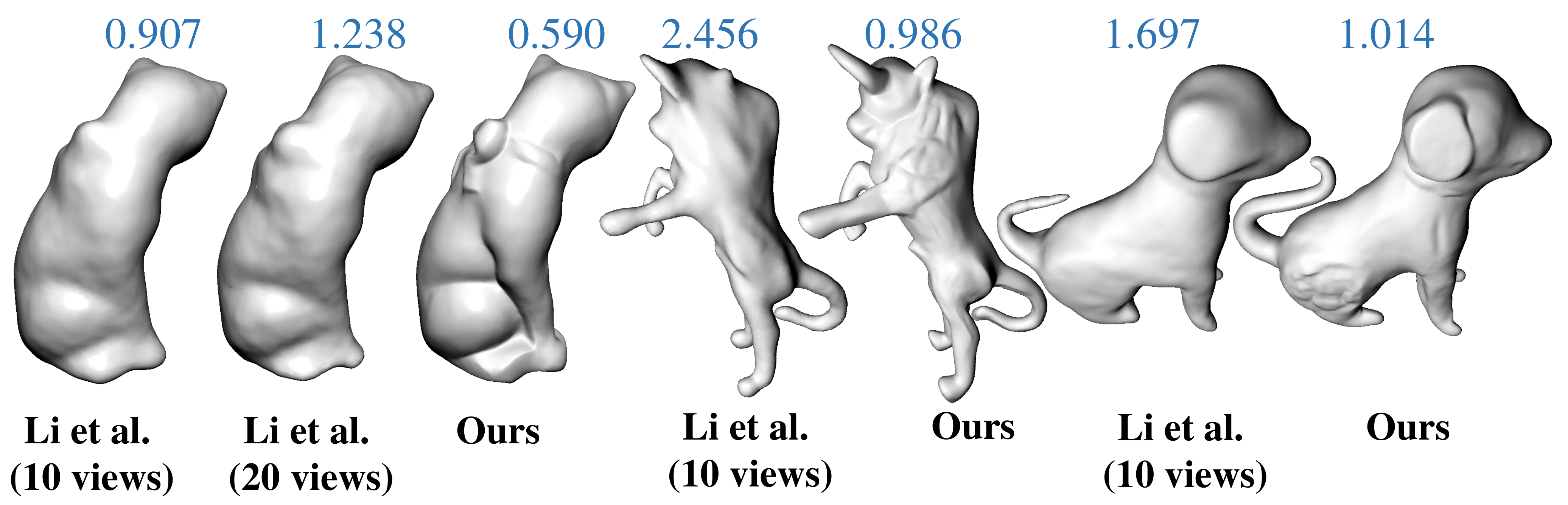}
\caption{Comparisons with Li et al.~\cite{li2020through}. The numbers indicate the chamfer distances between the mesh and the ground truth illustracted in Fig.~\ref{fig:comp_represenataion}.}
\label{fig:comparison_with_Li} 
\end{figure}

\begin{figure}[t!]
\centering
\small
\includegraphics[width=1.0\linewidth]{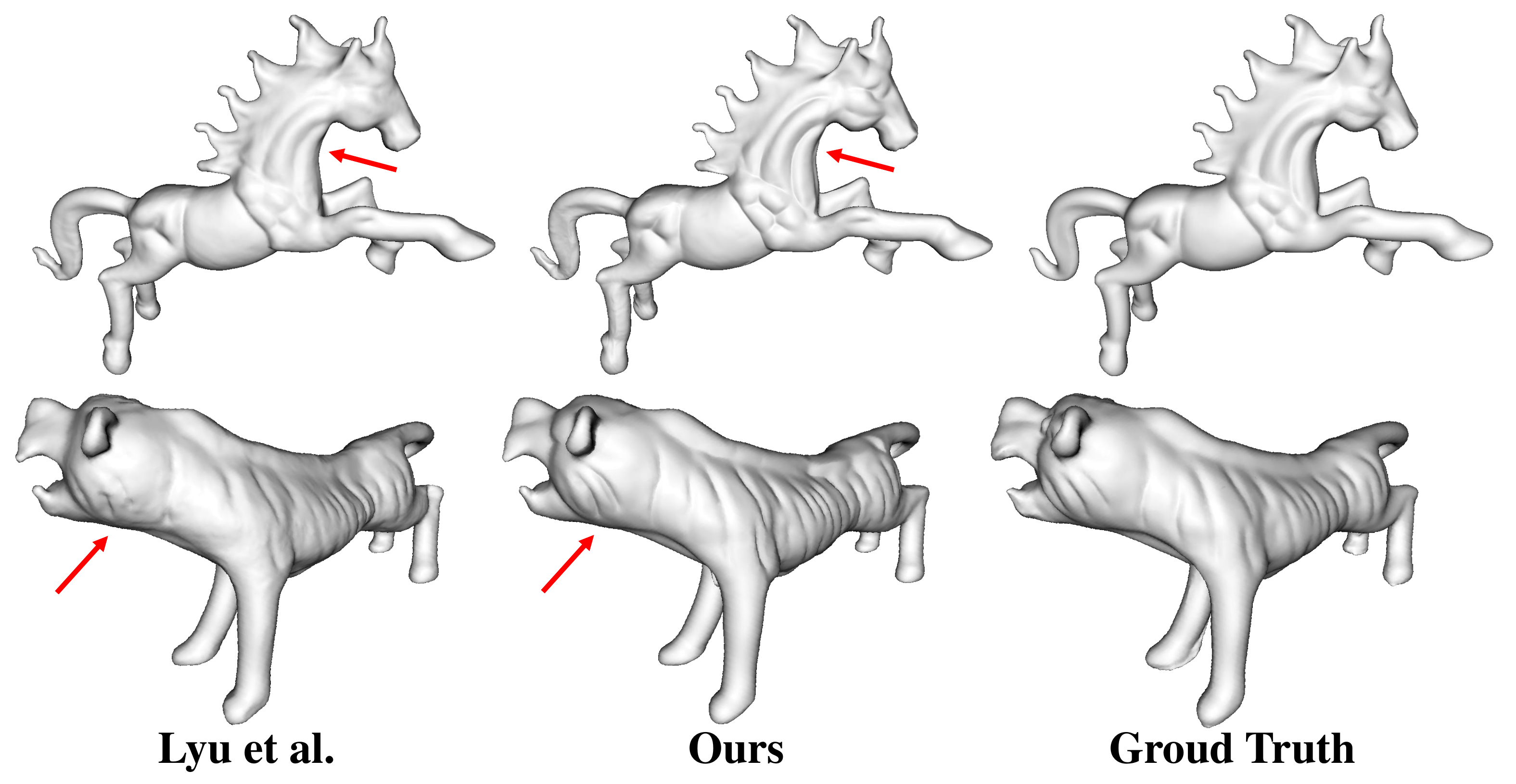}
\caption{Comparisons with Lyu et al.~\cite{lyu2020differentiable} using all 72 views as input. Our results preserve more surface details (indicated by the red arrow). The chamfer distances and mask differences are shown in Tab.~\ref{tab:comp_represenataion}.}
\label{fig:comparison_with_Lyu} 
\end{figure}

\jiamin{The method proposed by Li et al.~\cite{li2020through} needs to capture images in an environment large enough to meet the distant-illumination assumption for the environment map. Thus, we chose to evaluate this method using images obtained by rendering the ground-truth mesh illuminated by an environment map (no background geometry). Specifically, we used our ground truth mesh with one environment map to render multi-view images, corresponding object masks, and mirror sphere images, as shown in Fig. ~\ref{fig:li_etal_input_images}. The generated environmental map is shown in Fig. ~\ref{fig:li_etal_input_images}. A total of 36 images were rendered for visual hull reconstruction, and 10 of them were uniformly sampled as the input images to the network. As illustrated in Fig.~\ref{fig:comparison_with_Li}, this method produces smooth surfaces that lose detail near the object boundaries. Compared with the ground-truth mesh, our reconstructed model can preserve relatively high-precision details. \majorrev{Despite the results using 10 views, we also test 20 input views in our experiments. For the method proposed by Li et al.~\cite{li2020through}, we found that adding more views to the network did not improve the details of the reconstructed surface in this experiment, as shown in Fig. ~\ref{fig:comparison_with_Li}. Here, 10 or 20 views were used as the inputs of the network. The initial visual hull is calculated using all 36 images.}

For comparison with Lyu et al.~\cite{lyu2020differentiable}, as they only provide ray-pixel correspondences and masks in their dataset, we replace our RGB loss, corr loss, and ncorr loss with fraction loss~\cite{lyu2020differentiable}. As illustrated in Fig.~\ref{fig:comparison_with_Lyu}, our reconstructed model can preserve more surface detail. Qualitative and quantitative comparison results are shown in Fig. ~\ref{fig:comparison_with_Lyu} and Tab. ~\ref{tab:comparison_with_Lyu}. The ``mask diff'' is the mean L1 distance between rendered masks and the input masks. As shown in the fifth column of Table. ~\ref{tab:comparison_with_Lyu}, the ground-truth mesh is not precisely consistent with the input masks. This may be due to the DPT-5 developer coating used in the scanning. As shown in Tab.~\ref{tab:comparison_with_Lyu}, the chamfer distances of our results are slightly larger than those of this method. However, our mask difference is smaller, which means that our reconstructed model matches the input masks better. In this experiment, we added the mask difference measurement to demonstrate the advantage of our method because the silhouettes obtained by Lyu et al.~\cite{lyu2020differentiable} are much more accurate than those of the segmentation results.}

\begin{table}[t!] 
\caption{The chamfer distance and mask difference measurement for Fig.~\ref{fig:comparison_with_Lyu}.}
\centering
\resizebox{\linewidth}{!}{
\begin{tabular}{|c|c|c|c|c|}
\hline
{Object} & {Measurement}  & {Lyu et al.} & {Ours} & {Ground Truth} \\ 
\hline
{\multirow{2}{*}{\makecell{horse}}} & {chamfer dist$\downarrow$} & \textbf{0.003654}  & {0.004183} & {N/A}\\ 
\cline{2-5}
&
{mask diff$\downarrow$} & {0.003135}  & \textbf{0.002961} & {0.005236}\\
\hline
{\multirow{2}{*}{\makecell{tiger}}} & {chamfer dist$\downarrow$} & \textbf{0.005122}  & {0.005438} & {N/A}\\ 
\cline{2-5}
&
{mask diff$\downarrow$} & {0.004571}  & \textbf{0.004497} & {0.007975}\\
\hline
\end{tabular}
}
\label{tab:comparison_with_Lyu}
\end{table}

\subsubsection{Number of Clusters}
\begin{figure}[t!]
\centering
\includegraphics[width=0.85\linewidth]{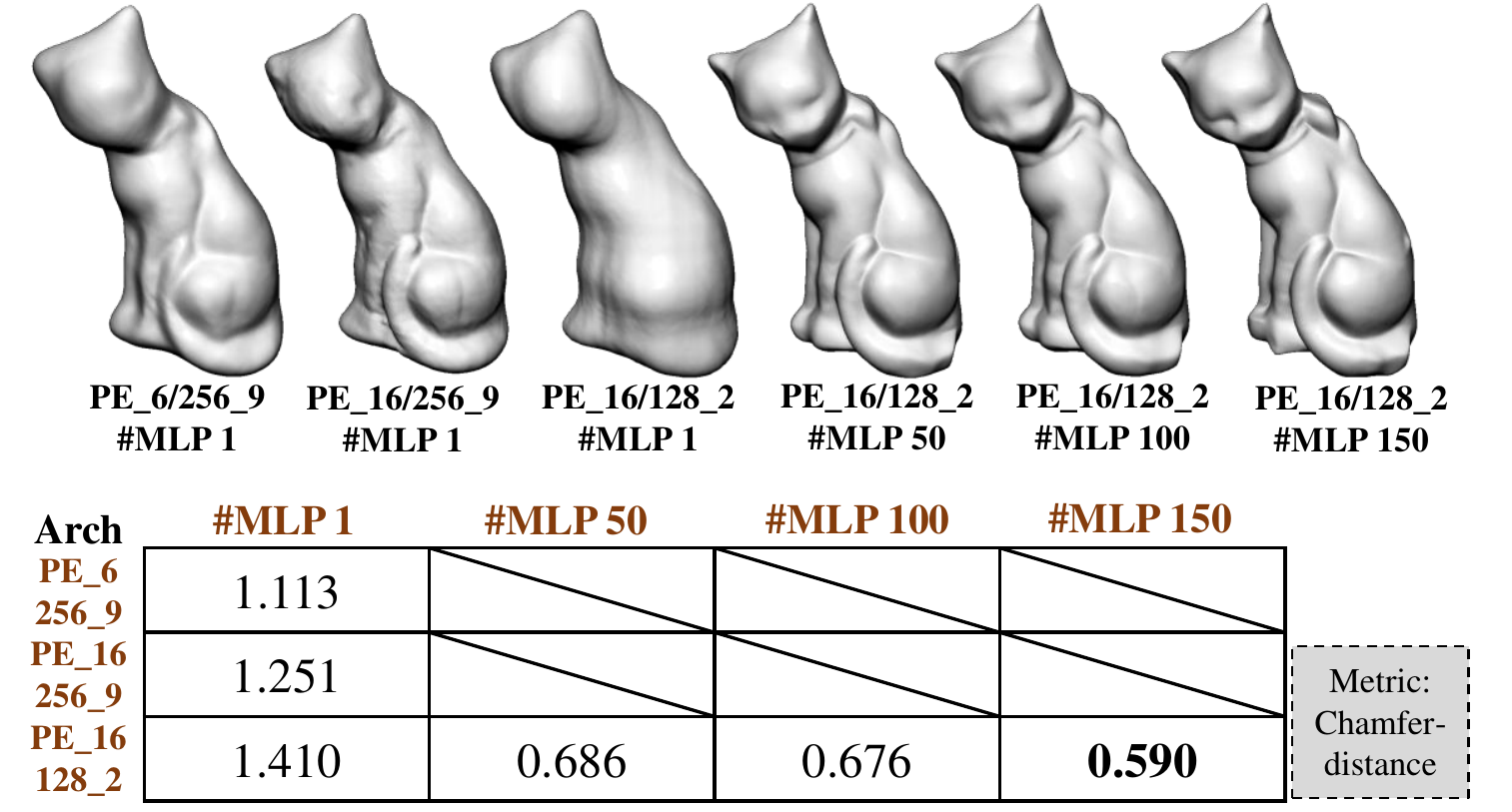}
\caption{The influence of MLP number (\#MLP) and the MLP architecture measured by the chamfer distance between reconstructed mesh and ground truth mesh. PE\_$k$/$d$\_$l$: positional encoding L=$k$, MLP with $l$ layer of $d$ dimension. All the results are obtained after training networks with 300 epochs. Back slash means the training cannot be performed due to computational cost.}
\label{fig:ablation_cluster_num} 
\end{figure}

We first performed ablation studies to evaluate the influence of the cluster number~(MLP number). Figure~\ref{fig:ablation_cluster_num} shows that our surface-based local MLP representation can obtain better results by increasing the number of clusters from 1, 50, 100, and 150. We also compared our local MLP representation to a global MLP with nine hidden layers of 256 dimensions. For the global MLP, we also test it with two different positional encoding settings, using the number of positional encoding frequencies L=6 and L=16. As illustrated in  Fig.~\ref{fig:ablation_cluster_num}, our multiple local MLP representation outperformed the single global MLP representation with a lower chamfer distance, despite the number of positional encoding frequencies. This experiment also shows that our local MLP representation can accelerate convergence: it can achieve a lower chamfer distance with the same number of epochs. We believe that this is because all the vertices share the MLP weights in the global MLP representation, which means VDF in  distant clusters will affect each other during training. 

\begin{figure}[t!]
\centering
\includegraphics[width=1.0\linewidth]{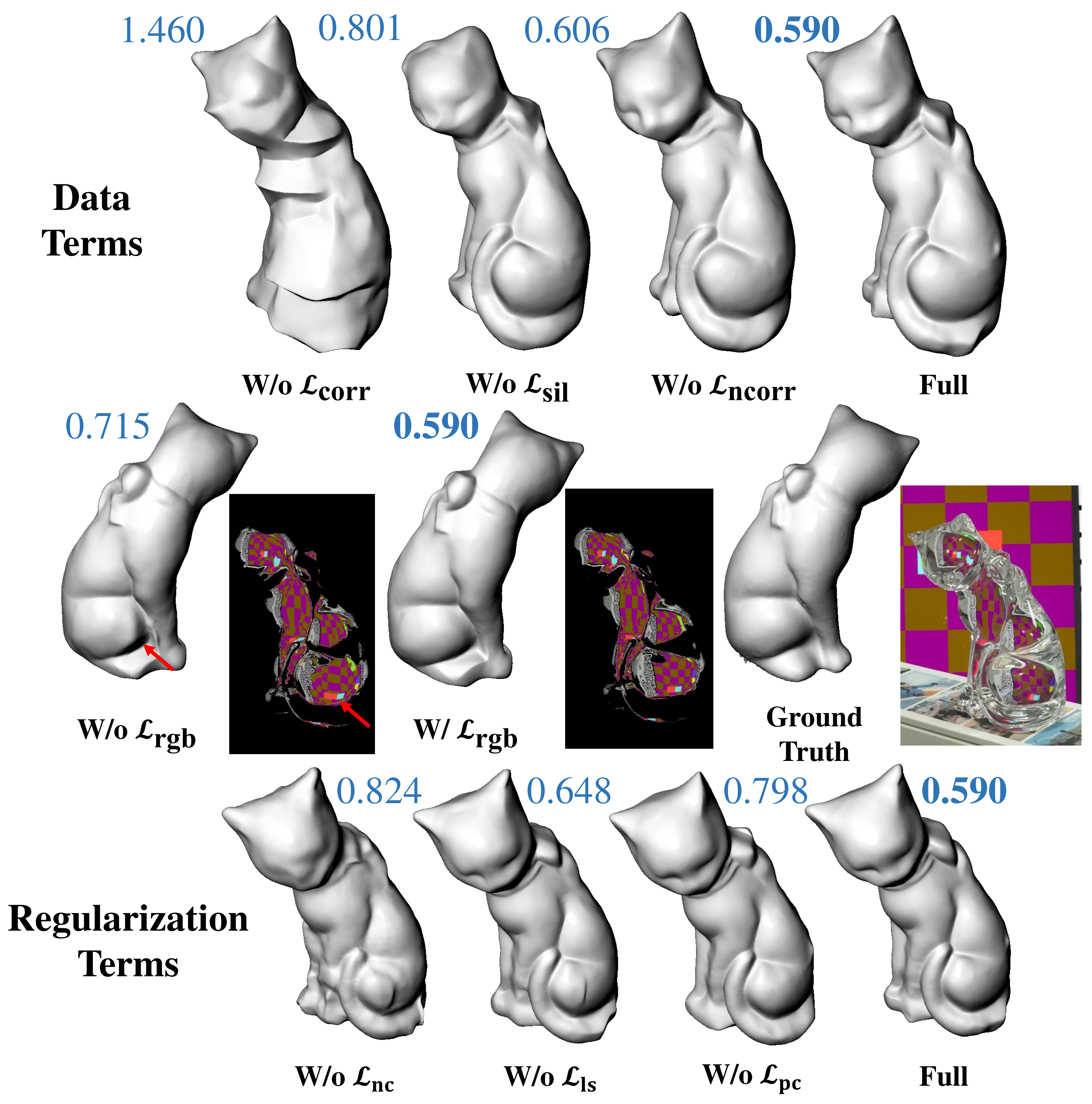}
\caption{Ablation study of loss terms on the cat object. The numbers indicate the chamfer distances between the reconstructed mesh and the ground truth mesh.}
\label{fig:ablation_loss} 
\end{figure}

\subsection{Ablation Study}

\majorrev{We remove each loss term individually to evaluate its impact on the reconstruction result of the cat object, as shown in Fig.~\ref{fig:ablation_loss}. Overall, the correspondence loss $\mathcal{L}_{\text{corr}}$, silhouette loss $\mathcal{L}_{\text{sil}}$, and normal consistency loss $\mathcal{L}_{\text{nc}}$ were the most important for our method. As shown in the first row in Fig. ~\ref{fig:ablation_loss}, $\mathcal{L}_{\text{corr}}$ is essential for our method to obtain surface details. $\mathcal{L}_{\text{sil}}$ can help preserve boundaries, such as the ears of the cat. The no-correspondence loss $\mathcal{L}_{\text{ncorr}}$ can also improve the result, which implies lower chamfer distances. However, because $\mathcal{L}_{\text{ncorr}}$ only constrains some intersection points outside the grid area, it cannot help too much for surface reconstruction.}

\majorrev{In the second row of Fig.~\ref{fig:ablation_loss}, we demonstrate the purpose of the RGB loss $\mathcal{L}_{\text{rgb}}$. The first two images were rendered using recursive ray tracing. The third image shows the corresponding patch of the input image. Our correspondence-based loss method can optimize the surface to move the refracted rays toward their corresponding cells. Then, as indicated
by the red arrow, RGB loss $\mathcal{L}_{\text{rgb}}$ can constrain the refracted rays near the cells by minimizing per-pixel color differences. By utilizing the gradient in each image's RGB space, $\mathcal{L}_{\text{rgb}}$ can fix small geometric inaccuracies and improve the surface details, as indicated by the first red arrow.}

\majorrev{We also test the purpose of each regularization term. The results are presented in the last row of Fig. ~\ref{fig:ablation_loss}. We can see that our method can obtain relatively good results without Laplacian smoothness loss $\mathcal{L}_{\text{ls}}$ and point cloud regularization loss $\mathcal{L}_{\text{pc}}$. However, $\mathcal{L}_{\text{ls}}$ can further improve surface smoothness, as shown in Fig. ~\ref{fig:ablation_loss}.  In our experiments, we found $\mathcal{L}_{\text{pc}}$ can improve the stability of our patch-based optimization.}

\section{Limitations and Future Work}
\label{sec:limitation}

Our environment-matting algorithm can only find correspondences, assuming that the transparent object has no intrinsic color. Consequently, our method cannot reconstruct colored transparent objects. Next, as the ray-cell correspondences are sparse, our method requires more views to reconstruct the surface and may miss some details, especially for surfaces with complex occlusions. Another limitation is that we need to manually annotate the object masks. In the future, it would be interesting to investigate how to integrate the variables for the color or other material properties of the transparent object to overcome the ``no intrinsic color'' limitation and how to extract accurate transparent object masks based on single-image EnvMatt. 





\section{Conclusions}
\label{sec:conclusion}

In this study, we developed a method to reconstruct 3D shapes of transparent objects from handheld captured images under natural light conditions. Our method comprises two components: a surface-based MLP representation that encodes the vertex displacement field based on the initial shape, surface optimization through differentiable rendering and EnvMatt. We used an \texttt{iPad} as a background to provide ray-cell correspondences, a simplified capture setting, to facilitate the optimization. Our method can produce high-quality reconstruction results with fine details under natural lighting conditions.



\subsection*{Acknowledgements}
We thank the anonymous reviewers for their constructive comments. Weiwei Xu is partially supported by the National Natural Science Foundation of China (No.61732016).

\subsection*{Declaration of competing interest}

The authors have no competing interests to declare that are relevant to the
content of this article.



\bibliographystyle{CVMbib}
\bibliography{reference}

\end{document}